\pdfoutput=1

\documentclass{article}

\usepackage{microtype}
\usepackage{graphicx}
\usepackage{subfigure}
\usepackage{booktabs} 
\usepackage{tabularray}
\usepackage{hyperref}

\usepackage[accepted]{icml2024}

\usepackage{amsmath}
\usepackage{amssymb}
\usepackage{mathtools}
\usepackage{amsthm}
\usepackage{xspace}
\usepackage{wrapfig}

\usepackage[capitalize,noabbrev]{cleveref}

\usepackage{amsmath,amsfonts,bm}

\def\eqref#1{equation~\ref{#1}}

\def\1{\bm{1}}

\def\vzero{{\bm{0}}}
\def\vone{{\bm{1}}}

\def\vdelta{{\bm{\delta}}}

\def\vd{{\bm{d}}}

\def\vh{{\bm{h}}}

\def\vp{{\bm{p}}}

\def\vr{{\bm{r}}}

\def\vt{{\bm{t}}}

\def\vx{{\bm{x}}}

\def\vz{{\bm{z}}}

\def\mC{{\bm{C}}}

\def\mH{{\bm{H}}}
\def\mI{{\bm{I}}}

\def\mS{{\bm{S}}}

\DeclareMathAlphabet{\mathsfit}{\encodingdefault}{\sfdefault}{m}{sl}
\SetMathAlphabet{\mathsfit}{bold}{\encodingdefault}{\sfdefault}{bx}{n}

\def\gB{{\mathcal{B}}}
\def\gC{{\mathcal{C}}}
\def\gD{{\mathcal{D}}}

\def\gH{{\mathcal{H}}}

\def\gN{{\mathcal{N}}}
\def\gO{{\mathcal{O}}}
\def\gP{{\mathcal{P}}}

\def\gR{{\mathcal{R}}}

\def\gX{{\mathcal{X}}}
\def\gY{{\mathcal{Y}}}

\def\sB{{\mathbb{B}}}

\def\sR{{\mathbb{R}}}

\newcommand{\E}{\mathbb{E}}

\newcommand{\mathfloor}[1]{\lfloor #1 \rfloor}
\newcommand{\mathceil}[1]{\lceil #1 \rceil}
\newcommand{\mathset}[1]{\left\{ #1 \right\}}
\newcommand{\prob}[1]{\mathrm{Pr}\left[ #1 \right]}
\newcommand{\quantile}[2]{\mathrm{Quant}\left(#1;#2\right)}
\newcommand{\quantileinline}[2]{\mathrm{Quant}(#1;#2)}
\newcommand{\invquantile}[2]{\mathrm{Quant}^{-1}\left(#1;#2\right)}
\theoremstyle{plain}
\newtheorem{theorem}{Theorem}[section]
\newtheorem{proposition}{Proposition}[section]

\theoremstyle{definition}
\newtheorem{definition}[theorem]{Definition}

\theoremstyle{remark}

\newcommand{\parbold}[1]{\textbf{#1}.\xspace}
\usepackage[inline]{enumitem}
\usepackage{pgf}

\usepackage{siunitx} 

\usepackage[textsize=tiny]{todonotes}

\newcommand{\method}{CAS\xspace}

\newcommand{\smoothf}{\xi}
\newcommand{\smooths}{\hat{s}}

\newcommand{\pertx}{\tilde{\vx}}
\newcommand{\perts}{\tilde{s}}
\newcommand{\pertq}{\tilde{q}}
\newcommand{\setgB}{\sB}
\newcommand{\Dcal}{\gD_{\mathrm{cal}}}
\newcommand{\conservativeSet}{\overline{\gC}}
\newcommand{\sHighE}{\overline{s}_+}

\newcommand{\sHigh}{\overline{s}}
\newcommand{\sLow}{\underline{s}}
\newcommand{\sHighMean}{\sHigh_{\mathrm{mean}}}

\newcommand{\sHighCdf}{\sHigh_{\mathrm{cdf}}}
\newcommand{\sLowCdf}{\sLow_{\mathrm{cdf}}}
\newcommand{\sLowCdfE}{\sLow_{\mathrm{cdf}+}}

\icmltitlerunning{Robust Yet Efficient Conformal Prediction Sets}

\begin{document}

\twocolumn[
\icmltitle{Robust Yet Efficient Conformal Prediction Sets}



\icmlsetsymbol{equal}{*}

\begin{icmlauthorlist}
\icmlauthor{Soroush H. Zargarbashi}{CISPA}
\icmlauthor{Mohammad Sadegh Akhondzadeh}{UOC}
\icmlauthor{Aleksandar Bojchevski}{UOC}
\end{icmlauthorlist}

\icmlaffiliation{CISPA}{CISPA Helmholtz Center for Information Security}
\icmlaffiliation{UOC}{University of Cologne}

\icmlcorrespondingauthor{Soroush H. Zargarbashi}{zargarbashi@cs.uni-koeln.de}

\icmlkeywords{Conformal prediction, randomized smoothing, adversarial robustness}

\vskip 0.3in
]



\printAffiliationsAndNotice{}  

\begin{abstract}
Conformal prediction (CP) can convert any model's output into prediction sets guaranteed to include the true label with any user-specified probability. However, same as the model itself, CP is vulnerable to adversarial test examples (evasion) and perturbed calibration data (poisoning).
We derive provably robust sets by bounding the worst-case change in conformity scores.
Our tighter bounds lead to more efficient sets.
We cover both continuous and discrete (sparse) data and our guarantees work both for evasion and poisoning attacks (on both features and labels).
\end{abstract}

\section{Introduction}
Uncertainty quantification (UQ) is crucial for deploying models, especially in safety-critical domains. The predicted probability is not a reliable source for UQ as it is often uncalibrated \citep{guo2017calibration}. 
Most methods do not provide any guarantees and require retraining or modifications in the model architecture \citep{abdar2021review}.
Instead, conformal prediction (CP) returns prediction \emph{sets} with a distribution-free guarantee to cover the true label.
It only requires black-box access to the model and assumes exchangeable data (a weaker assumption than i.i.d.).
This makes CP flexible -- we can apply it to image classification, segmentation \cite{angelopoulos2023conformal}, question answering \cite{Angelopoulos2022ConformalRC}, and node classification \cite{huang2023uncertainty}.

Most models suffer a significant performance drop when fed noisy or manipulated data, even for indistinguishable (label-preserving) perturbations \citep{silva2020opportunities}. 
Adversaries can exploit this vulnerability by perturbing the training data (poisoning) or the test data (evasion). CP's performance is also sensitive to the same attacks. One goal of the adversary is to break the guarantee -- reducing the probability to cover the true label by perturbing the test inputs (evasion) or poisoning the calibration data. In all settings, the perturbations are limited according to a threat model, e.g. a ball of a given radius around the clean input (see \autoref{sec:background}). 
Unlike heuristic defenses which are easily overcome by new attacks \citep{athalye2018obfuscated, mujkanovic2022defenses}, certificates provide worst-case guarantees that the prediction does not change.
How can we extend robustness certificates to conformal prediction sets?

Given calibration data and a score function $s:\gX\times\gY\mapsto\sR$ capturing conformity (agreement) between inputs and all potential labels, CP finds a calibrated threshold $q_\alpha$, and defines prediction sets $\gC_\alpha(\vx) = \{y : s(\vx, y) \ge q_\alpha\}$ that include all labels with scores above it.
CP guarantees that $\prob{y_\mathrm{true} \in \gC_\alpha(\vx)} \ge 1 - \alpha$ for a clean  $\vx$, exchangeable with the calibration data, and any user-specfied $\alpha$.
To certify robustness, we can define \emph{conservative} sets that ensure the coverage remains above $1-\alpha$ even under perturbation.

To this end,
\citet{gendler2021adversarially} 
leverage the fact that the randomly smoothed scores $ \E_{\vdelta \sim \gN(\vzero, \sigma^2 \mI)} [s(\vx + \vdelta, y)]$ change slowly around the input to compute an upper bound on the worst-case score.
Their \underline{r}andomly \underline{s}moothed \underline{c}onformal \underline{p}rediction (RSCP) method has 4 limitations: \begin{enumerate*}[label=(\roman*)]
    \item
    It considers only the mean
    of randomized scores
    resulting in a looser bound and thus larger sets;
    \item It only certifies evasion but not poisoning attacks;
    \item It only supports $L_2$-bounded perturbations of continuous data, ignoring discrete and sparse data such as graphs;
    \item It does not correct for finite-sample approximation errors.
\end{enumerate*} 
We address all of these limitations.

Our key insight is that we can use
the cumulative distribution (CDF) of smooth scores to obtain tighter upper bounds.
The resulting CDF-aware sets are smaller while maintaining the same robustness guarantee.
For continuous data we reuse \citet{kumar2020certifying}'s bound developed to certify confidence, while for discrete/graph data we extend the bounds of \citet{bojchevski2020efficient}.\footnote{Both of these methods do not provide sets or CP guarantees.}
We then propose an approach for finite sample correction. Different from \citet{anonymous2023provably}, we bound calibration points instead of test points. 
In addition to being significantly faster 
(especially for large datasets like ImageNet), 
our calibration-time algorithm also leads to smaller sets when correcting for finite samples.
\looseness=-1

%

Currently, there are no CP methods designed to handle poisoning. To fill this gap, we further derive provably robust sets that maintain worst-case coverage when either the features or the labels of the calibration set can be perturbed.
Moreover, the poisoning guarantee is independent of how the bound on conformity scores is derived.
Hence, our poisoning-aware and evasion-aware methods can be combined to provide robustness to both attacks simultaneously.


In short, we introduce \underline{C}DF-\underline{A}ware smoothed prediction \underline{S}ets (\method) that provably cover the true label under adversarial attacks.
For evasion, we show a consistent improvement on all metrics and datasets compared to RSCP.
Moreover, for the first time, we
additionally provide guarantees for poisoning, as well as discrete and sparse data.

\section{Background}
\label{sec:background}

\parbold{Conformal prediction}
Given a holdout calibration set $\gD_{\mathrm{cal}} = \mathset{(\vx_i, y_i)}_{i=1}^{n}$ exchangeably sampled from the data distribution (or a finite dataset) with labels unseen by the model (during training), and a user-specified coverage probability $1 - \alpha$, for any test point $\vx_{n+1}$, CP defines a prediction set $\gC_\alpha(\vx_{n+1})\subseteq \gY$ that is guaranteed to cover the true label $y_{n+1}$ with the predetermined probability.

\begin{theorem}[\citet{Vovk2005AlgorithmicLI}]
   \label{thrm:cp-base-theorem} 
    If $\Dcal=\{(\vx_i, y_i)\}_{i=1}^{n}$, and $(\vx_{n+1}, y_{n+1})$ are exchangeable,
     for any continuous score function $s: \gX \times \gY \mapsto \sR$ capturing the agreement between $\vx$, and $y$, and user-specified $\alpha \in (0, 1)$, the prediction set defined as $\gC_\alpha(\vx_{n+1}) = \mathset{ y: \ s(\vx_{n+1}, y) \ge q_\alpha}$ has coverage probability \begin{align}
        \label{eq:cp-guarantee}
         \prob{y_{n+1} \in \gC_\alpha(\vx_{n+1})} \ge 1 - \alpha
     \end{align}
     where $q_\alpha := \quantile{\alpha}{\{s(\vx_i, y_i)\}_{i = 1}^n}$ is the $\alpha$-quantile of the true scores in the calibration set.
\end{theorem}
This theorem was extended to graphs \citep{zargarbashi2023GraphCP, huang2023uncertainty} showing that the same guarantee holds for node classification. Although the coverage is guaranteed regardless of the choice of score function, a good choice is reflected in the size of the prediction sets (also called efficiency), the proportion of singleton sets covering the true label, and other metrics. 
A simple score function known as threshold prediction sets (TPS) directly considers the model's output $s(\vx, y) = \pi(\vx, y)$ where 
$\pi$ are the class probability (softmax) estimates \citep{Sadinle2018LeastAS}. TPS tends to over-cover easy examples and under-cover hard ones \citep{angelopoulos2021gentle}. This is remedied by the commonly used adaptive prediction sets (APS) score defined as $s(\vx, y):= - \left( \rho(\vx, y) + u \cdot \pi(\vx)_{y} \right)$. Here $\rho(\vx,y):= \sum_{c = 1}^{K} \pi(\vx)_c 1\left[\pi(\vx)_c > \pi(\vx)_y\right]$ is the sum of all classes predicted as more likely than $y$, and $u \in [0, 1]$ is a uniform random value that breaks the ties between different scores to allow exact $1 - \alpha$ coverage \citep{Romano2020ClassificationWV}. While we report our results on both scoring functions, our approach is orthogonal and hence applicable to any other choice (see \autoref{sec:conformal-more} for an extended introduction to CP).

\parbold{Adversarial attacks}
We define the threat model  -- the set of all possible perturbations the adversary can apply -- by a ball centered around a clean input $\vx$. For continuous $\vx$ we consider the $l_2$ ball of radius $r$ around the input $\gB_r(\vx) = \mathset{\pertx \in \gX: || \pertx - \vx ||_2 \le r}$.
For binary data, we define the ball w.r.t. the number of flipped bits:
$
  \gB_{r_a, r_d}(\vx) = \{\pertx \in \gX: \sum_{i=1}^d \1[\pertx_i = \vx_i - 1] \le r_d,$ 
  $\sum_{i=1}^d \1[\pertx_i = \vx_i + 1] \le r_a\}  
$ where $r_d$ and $r_a$ are the numbers of deleted and added bits respectively. This distinction accounts for sparsity as shown by \citet{bojchevski2020efficient}. We discuss categorical data in \autoref{sec:rand-smooth-bin}, extensions to other threat models are simple.

\parbold{Evasion attacks} For a given input $\vx$ and the model $f$, the adversary's usual goal is to find a perturbed input $\pertx$ such that $f(\pertx) \neq f(\vx)$ \citep{yuan2019adversarial, madry2017towards}. In CP, the goal changes to excluding the true label from the prediction set $\gC_\alpha(\pertx)$ which breaks the guarantee in \autoref{eq:cp-guarantee}. Here we assume that CP is calibrated with clean calibration points. \looseness=-1


\parbold{Poisoning attacks} The adversary can perturb the training data to e.g. decrease accuracy. However, since CP is model-agnostic, the guarantee holds regardless of the model's accuracy. Instead, here the goal of the adversary is to perturb the \emph{calibration} set in order to decrease the empirical coverage -- breaking the guarantee (see formal definition in \autoref{sec:robust-to-poisoning}).

\section{Robust Prediction Sets}
\label{sec:robust-cp}

\subsection{Robustness to Evasion Attacks}

\begin{definition}[Robust coverage] 
    The prediction sets $\gC_\alpha(\cdot)$ have adversarially robust $1 - \alpha$ coverage if for any $(\vx_{n+1}, y_{n+1})$ exchangeable with $\Dcal$
    \begin{align}
        \label{eq:robust-coverage}
        \prob{y_{n+1} \in \gC_\alpha(\pertx_{n+1}) \mid \pertx_{n+1} \in \gB(\vx_{n+1})} \ge 1-\alpha
    \end{align}
\end{definition}
where $\gB(\vx)$ can be the $l_2$ ball $\gB_r(\vx)$, the binary ball $\gB_{r_a, r_d}$, or any other threat model.
\citet{gendler2021adversarially} define a score $s_\mathrm{rscp}(\vx, y)=\Phi^{-1}(\E_{\vdelta \sim \gN(\vzero, \sigma^2 \mI)} [s(\vx + \vdelta, y)])$ based on Gaussian smoothing \citep{cohen2019certified} where $\Phi^{-1}(\cdot)$ is the inverse CDF of $\gN(0, 1)$.
Since the smooth score is bounded, $s_\mathrm{rscp}(\pertx, y) \le s_\mathrm{rscp}(\vx, y) + \frac{r}{\sigma}, \forall \pertx \in \gB_r(\vx)$ they shift the quantile $\underline{q_\alpha} = q_\alpha - \frac{r}{\sigma}$ to ensure robustness. 
Instead of shifting the quantile we directly bound the conformal scores which is a slight generalization. 

\begin{proposition}
\label{thrm:conservative-guarantee}
    Define $\sHigh(\vx, y)$ as the upper bound for $\mathset{s(\pertx, y): \pertx \in \gB(\vx)}$. With $q_\alpha$ as the $\alpha$-quantile of the true (clean) calibration scores, let $\conservativeSet_\alpha(\vx) = \mathset{y: \sHigh(\vx, y) \ge q_\alpha}$. For all $\pertx_{n+1} \in \gB(\vx_{n+1})$, if $(\vx_{n+1}, y_{n+1})$ is exchangeable with $\Dcal$ then we have
    $
       \prob{y_{n+1} \in \conservativeSet_\alpha(\pertx_{n+1})} \ge 1 - \alpha
   $.
\end{proposition}
All omitted proofs are in \autoref{sec:proofs}.
We summarize our notation in \autoref{sec:notations}.
In short, the conservative set for any $\pertx \in \gB(\vx)$ includes the labels of the vanilla prediction set for $\vx$. Thus, the coverage guarantee also applies for the perturbed points. 

RSCP is a special case with $\sHigh(\vx, y) = s_\mathrm{rscp}(\vx, y) + \frac{r}{\sigma}$.
We can equivalently rewrite RSCP
as an upper bound on $\E[s(\cdot, \cdot)]$
instead of $\Phi^{-1}(\E[s(\cdot, \cdot)])$
which matches the bound from \citet{kumar2020certifying} (see \autoref{sec:comparison-rscp}).
In \autoref{sec:efficienct-bounds} we significantly improve the bound using the CDF. Tighter bounds result in smaller (more efficient) sets. 

\subsection{Robustness to Feature Poisoning Attacks}
\label{sec:robust-to-poisoning}
We assume that the adversary can modify at most $k$ instances, $0 \le k \le n = |\Dcal|$, whose features can be perturbed in a (continuous or discrete) ball $\gB$ around the clean features.
We define the threat model at dataset-level:
\begin{align*}
\setgB_{k, \gB}(\gD) = \{\tilde\gD: \tilde\gD=\{(\pertx_i, y_i): (\vx_i, y_i) \in \gD, \\
       \pertx_i \in \gB(\vx_i), \sum_{j = 1}^n \1[\pertx_j \neq \vx_j] \le k\}\}
\end{align*}
Let $q_\alpha$ be the $\alpha$-quantile of the clean calibration scores. To decrease coverage the adversary aims to find a perturbed calibration set $\tilde{\gD}_\mathrm{cal} \in \setgB_{k, \gB}(\Dcal)$ that moves the quantile 
$\pertq_\alpha = \quantileinline{\alpha}{\tilde{\gD}_\mathrm{cal}}$ 
as right as possible compared to $q_\alpha$.\footnote{Our setup works with conformity score capturing the agreement between $\vx$ and $y$. With a non-conformity score, the goal is to equivalently shift the quantile to the left (see \autoref{sec:conformal-more}).} This shift increases the probability of rejecting true labels, resulting in a lower coverage. Namely, for $\tilde\alpha = \mathrm{Quant}^{-1}(\pertq_\alpha; \Dcal)$, the quantile inverse of the poisoned threshold $\tilde{q}$ w.r.t. the clean calibration set, the poisoned calibration set results in near $1 - \tilde\alpha$ coverage where by definition $1 - \tilde\alpha \le 1 - \alpha$. Given a potentially poisoned calibration set $\tilde{\gD}_\mathrm{cal}$ we certify the prediction sets via the following optimization problem:
\begin{equation}
    \label{eq:poison-problem}
    \begin{aligned}
        \underline{q_\alpha} = & \min_{\vz_i \in \gX} \  \quantile{\alpha}{\{s(\vz_i, y_i)\}_{i=1}^{n}} \\
        \text{s.t.} \quad  &  \forall (\pertx_i, y_i) \in \tilde\gD_\mathrm{cal}: \vz_i \in \gB(\pertx_i) \\ & \sum_{i\le n}\1[\vz_i \neq \pertx_i] \le k
    \end{aligned}
\end{equation}
The problem in \autoref{eq:poison-problem} finds the most conservative quantile $\underline{q_\alpha}$ and it holds that $\underline{q_\alpha} \leq q_\alpha$ 
since for any perturbed $\tilde\gD_\mathrm{cal}$ by definition it holds $\Dcal\in \setgB_{k, \gB}(\tilde\gD_\mathrm{cal})$. 
We show that the minimizer of problem \autoref{eq:poison-problem} certifies \emph{at least} $1 - \alpha$ coverage.

\begin{proposition}
    \label{thrm:poisoning}
    Let $\underline{q_\alpha}$ to be the solution to the optimization problem in \autoref{eq:poison-problem}. With the conservative prediction sets
    \begin{align}
    \label{eq:conservative-poisoning}
\conservativeSet_\alpha(\vx_{n+1}) = \mathset{y_i: s(\vx_{n+1}, y_i) \ge \underline{q_\alpha}}
    \end{align}
    for any $(\vx_{n+1}, y_{n+1})$ exchangeable with (clean) $\Dcal$ we have $\prob{y_{n+1} \in \conservativeSet_\alpha(\vx_{n+1})} \ge 1 - \alpha$.
\end{proposition}

With access to lower and upper bounds on the adversarial scores we can change the constraint $\vz_i \in \gB(\pertx_i)$ in \autoref{eq:poison-problem} to $z_i \in [\sLow(\pertx_i, y_i), \sHigh(\pertx_i, y_i)]$ where $z_i \in \sR$ is a scalar variable, and solve the relaxed problem. We describe in \autoref{sec:efficienct-bounds} how to obtain such bounds using randomized smoothing which we can use in both \autoref{thrm:conservative-guarantee} and \autoref{thrm:poisoning}.
Regardless of how we solve \autoref{eq:poison-problem}, as long as it finds a $\underline{q_\alpha} \le q_\alpha$ conditional on the clean $\Dcal$ the guarantee holds.


\subsection{Robustness to Label Poisoning Attacks}
\label{sec:label-poisoning}
In the label poisoning setup, the adversary can flip the labels of at most $k$ datapoints in the calibration set, again aiming to shift the quantile to the right. As before, we can find the most conservative quantile by solving the problem:
\begin{equation}
    \label{eq:label-poison-problem}
    \begin{aligned}
        \underline{q_\alpha}= &  \min_{z_i \in \gY} \quantile{\alpha}{\mathset{s(\vx_i, z_i) : (\vx_i, \tilde{y}_i) \in \tilde{\gD}_\mathrm{cal}}} \\
        \text{s.t.} \quad &  \sum_{i \le n}\1[z_i \neq \tilde{y}_i] \le k
    \end{aligned}
\end{equation}
Similar to \autoref{sec:robust-to-poisoning}, since $\underline{q_\alpha} \leq q_\alpha$, prediction sets defined as in \autoref{eq:conservative-poisoning} maintain $\ge 1 - \alpha$ coverage even under worst-case label perturbation.
We can solve both problems (\autoref{eq:poison-problem} and \autoref{eq:label-poison-problem}) by writing them as mixed-interger linear programs (MILPs). We present the technical details in \autoref{sec:technical-poisoning}.

Interestingly, our evasion-aware sets can easily be combined with our poisoning-aware threshold to obtain prediction sets that are robust to both types of attacks. Similarly, we can easily combine the feature and label poisoning constraints in a single problem. We discuss these extensions in \autoref{sec:combined-robustness}.

\section{Randomized Smoothing Bounds}
\label{sec:efficienct-bounds}
To instantiate the conservative sets $\conservativeSet_\alpha(\cdot)$ defined in \autoref{sec:robust-cp} we need bounds on the worst-case change in conformity scores under perturbation. There is a rich literature on robustness certificates for standard classification \citep{li2023sok} that we can lean on, since they often need to compute similar bounds as a byproduct. We focus on methods based on the randomized smoothing framework \citep{cohen2019certified} given their high flexibility and black-box nature. This couples well with the flexibility of CP, ensuring that our final robust CP method can be broadly applied.

\parbold{Smooth scores}
A smoothing scheme $\smoothf:\gX\mapsto\gX$ is a function that maps the input $\vx$ to a nearby random point. Given an arbitrary score $s(\cdot, \cdot)$,
we compute the expected (smooth) conformal scores as $\hat{s}(\vx, y) := \E[s(\smoothf(\vx), y)]$. 
Following \citet{cohen2019certified} for Gaussian smoothing, we add isotropic noise where the scale $\sigma^2$ determines the amount of smoothing $\hat{s}(\vx, y) = \E_{\vdelta \sim \gN(\vzero, \sigma^2 \mI)} [s(\vx + \vdelta, y)]$.
For binary data, we use sparse smoothing \citep{bojchevski2020efficient} and flip zeros and ones with probabilities $p_0$ and $p_1$ respectively: 
$\hat{s}(\vx, y) = \E [s(\vx \oplus \vdelta, y)]$, where $\oplus$ is the XOR and each entry 
$\vdelta[i] \sim \textrm{Bernoulli }(p=p_{\vx[i]})$.
See \autoref{sec:rand-smooth-bin} for more details. Our approach works with other smoothing schemes such as uniform noise for $l_1$ threat models \citep{levine2021improved}, but we focus on these two due to their popularity.
Gaussian smoothing preserves exchangeability \citep{gendler2021adversarially}. Similar argument applies to sparse smoothing and other methods that are symmetric w.r.t. $\vx_{n+1}$ and $\Dcal$.

The goal is to bound the smooth score $\hat{s}(\pertx, y)$ of any adversarial $\pertx \in \gB(\vx)$. Since the base score function $s(\cdot, \cdot)$ often depends on a complex model such as a neural network, even computing the expected score $\hat{s}(\cdot, \cdot)$ is challenging, let alone finding the worst-case $\pertx$. Therefore, we follow the general recipe of relaxing the problem by searching over the space of all possible score functions $h(\cdot, \cdot) \in \gH$. We focus on upper bounds, but the entire discussion equivalently applies to lower bounds by switching from $\max$ to $\min$. By definition we have $s(\cdot, \cdot) \in \gH$, therefore it holds that:
\begin{align}
    \label{eq:upper_bound_trivial}
    \max_{\pertx \in \gB(\vx)} \E[s(\smoothf(\pertx), y)] \leq \max_{\pertx \in \gB(\vx), h \in \gH} \E[h(\smoothf(\pertx), y)]
\end{align}

The solution to \autoref{eq:upper_bound_trivial} is trivial unless we add additional constraints to the functions $h(\cdot, \cdot) \in \gH$ that capture information about the actual score function $s(\cdot, \cdot)$. The tightness of the resulting bound is directly controlled by the constraints. 
First, we describe a baseline bound that only captures information about the mean of $s(\cdot, \cdot)$. This is exactly the bound used by RSCP. Then, we describe a second bound that leverages information about the entire distribution of scores via the CDF. In both cases, we only need black-box access to the score function and the underlying classifier, and we assume that $s(\cdot, \cdot) \in [a, b]$ is bounded (w.l.o.g.  $a=0, b=1$).


\parbold{Canonical view} It turns out that for both Gaussian and sparse smoothing it is sufficient to derive a so-called point-wise bound for a given $(\vx, \pertx)$ pair since it can be shown that the maximum in \autoref{eq:upper_bound_trivial} is always attained at a canonical $\pertx$ which is on the sphere of the respective ball. Namely, for the continuous $\gB_r(\vx)$ we have the canonical vectors $\vx=\vzero, \pertx=[r, 0, 0, \dots]$ that completely specify the problem. For the binary $\gB_{r_a, r_d}$ we have the canonical $\vx=[1, \dots, 1, 0, \dots, 0]$ and $\pertx=\vone - \vx$ where $\|\vx\|_0=r_d$ and $\|\pertx\|_0=r_a$. Intuitively, the reason is due to the symmetry of the smoothing distributions and the balls (see \autoref{sec:rand-smooth-bin}). 

\parbold{Baseline bound} A straightforward approach only incorporates the expected smoothed score (mean) for the given input $\vx$. Let $p = \E[s(\smoothf(\vx), y)]$ for simplicity. With $\pertx \in \gB(\vx)$ the baseline upper-bound for $\hat{s}(\pertx, y)=\E[s(\smoothf(\pertx), y)]$ is determined by the following problem:
\begin{equation}
    \label{eq:certificate-smooth}
    \begin{aligned}
        \sHighMean(\vx, y) &= \max_{h \in \gH} \quad  \E[h(\smoothf(\pertx), y)] \\
        &\mathrm{s.t.} \quad  \E[h(\smoothf(\vx), y)] = p
    \end{aligned}
\end{equation}
This bound discards a lot of information about the distribution of scores around the given $\vx$. To remedy this, we incorporate the information from the CDF of the scores. 

\parbold{CDF-based bound} 
Let $a = b_1 < b_2 \le \dots \le b_{m-1} < b_m = b$ be $m$ real numbers that partition the output space. Let $p_{i} = \prob{s(\smoothf(x), y) \le b_i}$. We define the problem: 
\begin{equation}
    \label{eq:certificate-smooth-cdf}
    \begin{aligned}
        \sHighCdf(\pertx, y) &= \max_{h \in \gH} \quad  \E[h(\smoothf(\pertx), y)]\\
        &\mathrm{s.t.}\quad \forall b_i: \quad  \prob{h(\smoothf(\vx), y) \le b_i} = p_i
    \end{aligned}
\end{equation}
%
The key insight for solving \autoref{eq:certificate-smooth-cdf} is to upper bound the mean of $h$ via the CDF. 
Intuitively, we compute the probability of each bin $[b_j, b_{j+1}]$ and choose the upper end of the bin to get an upper bound. This can be rewritten in terms of the CDF. 
Let $F_h(b_j) = \prob{h(\vx,y) \le b_{j}}$, for any function $h$
\begin{equation}
    \label{eq:anderson}
    \begin{aligned}
        \E[h(\vx)] \le & \sum_{j=2}^{m} b_{j} \cdot [(F_h(b_{j})  - F_h(b_{j-1})]  \\
        = & \ b_m - \sum_{j=2}^{m-1} F_h(b_{j}) \cdot (b_{j+1} - b_{j})
    \end{aligned}
\end{equation}
Next, we show how to solve both problems for the two different smoothing schemes. For Gaussian smoothing, both problems in \autoref{eq:certificate-smooth} and \autoref{eq:certificate-smooth-cdf} have closed-form solutions as shown by \citet{kumar2020certifying}. For sparse smoothing, \citet{bojchevski2020efficient} provides an efficient algorithm to solve \autoref{eq:certificate-smooth}. We extend their approach to also solve \autoref{eq:certificate-smooth-cdf} which is a novel contribution of potentially independent interest, e.g. to certify graph neural networks with regression tasks. \looseness=-1

In practice, $\sHighCdf$ is tighter than $\sHighMean$, and the improvement depends on the distribution of random scores. While we can easily combine both mean and CDF constraints to get a provably tighter bound, we focus only on CDF constraints.

\parbold{Bounds for Gaussian smoothing}
For any perturbed   $\pertx$ with $|| \pertx - \vx ||_2 \le r$ we have the baseline bound $\hat{s}(\pertx, y) \le \sHighMean(\vx, y) = \Phi_\sigma\left(\Phi^{-1}_\sigma\left( p \right) + r\right)$ where $\Phi_\sigma$ is the CDF of $\gN(0, \sigma^2)$ and $p=\E_{\vdelta \sim \gN(\vzero, \sigma^2\mI)}[s(\vx + \vdelta, y)]$ is the clean expected score. We can get the lower bound by flipping the sign of $r$. 
The CDF bound is $\hat{s}(\pertx, y) \le \sHighCdf(\vx, y)$ with
\begin{align}
    \label{eq:gaussian_cdf}
    \sHighCdf = b_m  -  \sum_{j = 2}^{m-1} \Phi_\sigma\left( \Phi^{-1}_\sigma({p_j})-r \right)(b_{j+1} - b_{j})
\end{align}
where $p_j = \mathrm{Pr}_{\vdelta \sim \gN(\vzero, \sigma^2\mI)}[s(\vx +\vdelta, y) \le b_j]$.
The corresponding lower bound and derivations are in \autoref{sec:cdf-bounds}.

\parbold{Bounds for sparse smoothing}
To solve both optimization problems, we apply the same approach as \citet{bojchevski2020efficient}, dividing the input space into regions of constant likelihood ratio $\gX = \cup_{i}^{I}\gR_i$ where $\gR_i = \mathset{\vz: \prob{\smoothf(\vx) = \vz} / \prob{\smoothf(\pertx) = \vz}=c_i}$.
For the mean variant, we greedily distribute the $p$ mass to each region (from the highest to the lowest ratio) until the constraint is satisfied.
For the CDF variant, we instead distribute the $p_j$ masses in each region and each bin $[b_j, b_{j+1}]$. Technical details, including the linear programming formulations, are in \autoref{sec:rand-smooth-bin}. The runtime complexity scales linearly with the number of regions which is $I=r_a+r_d+1$. We provide an efficient algorithm that runs in less than a few milliseconds.

\parbold{Clean vs. observed input}
In the discussion we refer to a clean $\vx$ and a perturbed $\pertx \in \gB(\vx)$.
In practice, we do not know whether the \emph{observed} input $\vx'$ is clean or perturbed.
However,  since the $l_2$-ball is symmetric, if $\vx' \in \gB_r(\vx)$ then also $\vx \in \gB_r(\vx')$.
Thus, computing an upper bound for any observed $\vx'$ in the threat model yields a valid upper bound for the clean $\vx$, $\hat{s}(\vx) \leq \sHigh(\vx')$. That is, we do not assume that the clean input is given at test time. 
For sparse data $\vx' \in \gB_{r_a, r_d}(\vx) \implies \vx \in \gB_{r_d, r_a}(\vx')$, so we need to switch $r_a$ and $r_d$ when computing the certificate.
Similar conclusions apply for an observed and potentially perturbed $\Dcal'$ since the clean $\Dcal\in \setgB_{k, \gB}(\gD_\mathrm{cal}')$ for any $\Dcal'\in \setgB_{k, \gB}(\gD_\mathrm{cal})$.
This detail is not important for standard certificates since they only certify that the prediction does not change.



\section{\method: CDF-Aware Sets}
\label{sec:method}

We use the CDF-based bounds to obtain conservative prediction sets for evasion and conservative thresholds for poisoning attacks. 
We summarize our approach with the pseudo-code in \autoref{alg:evasion} that works with any score function\footnote{Our code and experiments are in \href{https://github.com/soroushzargar/CAS}{the github repository \texttt{soroushzargar/CAS}}.}. \looseness=-1

\begin{algorithm}
\caption{CDF-Aware Sets (\method, Evasion)}
\label{alg:evasion}
\begin{algorithmic} 
        \STATE $q_\alpha = \quantile{\alpha}{\{\hat{s}(\vx, y)}_{(\vx, y) \in\gD_\mathrm{cal}}\}$  \text{\ $\triangleright$ \ Clean quantile}
        \STATE Compute $\sHighCdf(\vx, y)$, e.g. with \autoref{eq:gaussian_cdf} \text{\ $\triangleright$ \ Upper bound}
        \STATE Return $\conservativeSet_\alpha=\mathset{y: \sHighCdf(\vx, y) \ge q_\alpha}$ \text{\ $\triangleright$ \ Conservative set}
\end{algorithmic}
\end{algorithm} 

\parbold{Calibration-time variant}
For evasion we need to compute $\sHigh(\vx, y)$ via solving \autoref{eq:certificate-smooth-cdf} (or \autoref{eq:certificate-smooth}) for each test point and each class. This can be computationally costly if we have many classes (e.g. \texttt{ImageNet} has 1000) at deployment. 
We define an alternative approach that instead needs only a \emph{lower bound} $\sLow(\vx, y)$ for each $\vx \in \Dcal$ and the true $y$.
The key insight is that we can directly compare the smooth test score $\hat{s}(\pertx_{n+1}, y)$ against a conservative (lower) quantile.

\begin{proposition}
    \label{thrm:worst-case-coverage}
    For $\pertx_{n+1} \in \gB(\vx_{n+1})$ and $(\vx_{n+1}, y_{n+1})$ exchangeable with $\Dcal$, define
    \begin{equation}
    \label{eq:cal-time-conservative}
        \underline{q_\alpha} = \quantile{\alpha}{\mathset{\sLow(\vx_i,y_i): (\vx_i,y_i)\in\Dcal}}
    \end{equation}
    For prediction sets $\conservativeSet_\alpha(\pertx_{n+1}) = \{y: \hat{s}(\pertx_{n+1}, y) \ge \underline{q_\alpha}\}$
    we have $\Pr[{y_{n+1} \in \conservativeSet_\alpha(\pertx_{n+1})}] \ge 1 - \alpha$.
    Moreover, the vanilla CP covers the true label with probability $\ge 1 - \beta$ for \looseness=-1 \begin{equation}
        \label{eq:lower-bound-certified-coverage}
        \beta = \invquantile{q_\alpha}{\mathset{\sLow(\vx_i,y_i): (\vx_i,y_i)\in\Dcal}}
    \end{equation}
    and $\invquantile{t}{A} = \min\mathset{\tau': \quantile{\tau'}{A} \ge t}$.
\end{proposition}

With \autoref{thrm:worst-case-coverage} we need only $|\Dcal|$ certified bounds as a pre-processing step.
At test time we directly plug in $\hat{s}(\pertx_{n+1}, y)$ and not its upper bound. Since $|\Dcal|$ is often significantly smaller than the test set the computational savings are substantial (see \autoref{tab:comparison_runtime}). 
With \autoref{eq:lower-bound-certified-coverage} we can compute a lower bound on the coverage of vanilla (non-robust) CP under perturbation, where by definition $1-\beta \le 1-\alpha$. This is a generalization of Theorem 2 in \citet{gendler2021adversarially}.

\parbold{Poisoning}
For poisoning attacks we simply use the conservative threshold $\underline{q_\alpha}$ from \autoref{eq:poison-problem} or \autoref{eq:label-poison-problem} where we use the CDF-bounds in the constraints (see \autoref{sec:robust-to-poisoning}). 
If the test examples are assumed clean we return $\conservativeSet_\alpha=\{y: \hat{s}(\vx, y) \ge \underline{q_\alpha}\}$. Since robustness to evasion and poisoning are independent, we can achieve simultaneous robustness to both evasion and poisoning via $\conservativeSet_\alpha=\{y: \sHighCdf(\vx, y) \ge \underline{q_\alpha}\}$.

To solve the two poisoning optimization problems we rewrite them as mixed-integer linear programs and solve them with an off-the-shelf solver. We only need $2\cdot|\Dcal|$ binary variables for \autoref{eq:poison-problem} and $|\Dcal|\times|\gY|$ binary variables for \autoref{eq:label-poison-problem}. See \autoref{sec:technical-poisoning} for technical details. Since the calibration set is relatively small we can solve the MILPs in just a few minutes. Thus, our guarantees are practically feasible.

\begin{figure*}[t!]
    \centering
    \input{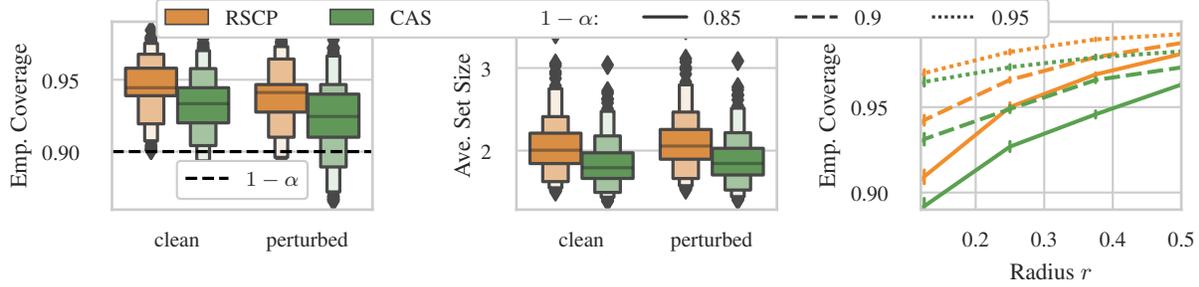}
    \caption{Empirical coverage [left] and average set size [middle] of RSCP and \method for clean and perturbed data. All sets are certified robust up to radius $r=0.125$. [Right] Empirical coverage for different certified radii (on clean data). All results are for \texttt{CIFAR-10} with Gaussian smoothing ($\sigma=0.25$).
    \method is less conservative since it is closer to the nominal $1-\alpha$, and has smaller sets.
    }
    \label{fig:attacked-data}
\end{figure*}


\section{Finite Sample Correction}
\label{sec:sample-correction}
Solving \autoref{eq:certificate-smooth}, or \autoref{eq:certificate-smooth-cdf} requires the true mean or CDF. Since exact computation is intractable, we use Monte-Carlo (MC) samples. To ensure a valid certificate, we bound the exact statistics via concentration inequalities. The resulting confidence intervals are valid together with adjustable $1 - \eta$ probability. To account for this we calibrate with $\alpha'=\alpha - \eta$ so that the final sets still have $1-\alpha$ coverage (see \autoref{sec:monte-carlo}).
RSCP did not include such finite-sample correction, and the resulting sets are only asymptotically valid without it.

\citet{anonymous2023provably} incorporates the correction directly in the conformity scores, leveraging exchangeability between MC-estimated calibration scores and clean test scores. We discuss this in  \autoref{sec:monte-carlo} and propose another approach built on \autoref{thrm:worst-case-coverage}.
Our correction results in smaller sets for \method with the same guarantee; and similar results for RSCP  (see \autoref{sec:experiments}). \looseness=-1
\begin{proposition}
\label{thrm:conformal-mc}
    Let $\sLowCdfE(\vx_{i}, y_i) \le \sLowCdf(\vx_i, y_i)$ hold with $1 - \eta/(2|\Dcal|)$ probability for each $(\vx_i, y_i)\in\Dcal$, and $\smooths_+(\pertx_{n+1}, y_{n+1}) \ge \smooths(\pertx_{n+1}, y_{n+1})$ hold with $1 - \eta/(2|\gY|)$ probability. Define the conservative $\underline{q}_{\alpha+}=\quantile{\alpha - \eta}{\mathset{\sLowCdfE(\vx_{i}, y_i):(\vx_{i}, y_i)\in\Dcal}}$ and $\conservativeSet_{\alpha+}(\vx_{n+1}) = \{y: \smooths_+(\vx_{n+1},y) \ge \underline{q}_{\alpha+}\}$. Then
    \begin{align}
        \Pr[y_{n+1} \in \conservativeSet_{\alpha+}(\pertx_{n+1})] \ge 1 - \alpha
    \end{align}
\end{proposition}
We compute $\sLowCdfE(\vx_{i}, y_i)$ by solving the minimization variant of \autoref{eq:certificate-smooth-cdf} with CDF error correction through the Dvoretzky–Kiefer–Wolfowitz inequality \cite{dvoretzky1956asymptotic}. We define $\smooths_+(\pertx_{n+1}, y) = \frac{1}{n_s}\sum^{n_s} s(\smoothf(\vx_{n+1}), y) + \epsilon$ where $\epsilon$ is the error given by the Bernstein confidence interval. In short, we divide the $\eta$ budget between $|\Dcal| + |\gY|$ estimates. This divides between all calibration scores (only for the true class), and $|\gY|$ classes for the test input.

The corrected CP in \citep{anonymous2023provably} compares $q_{\alpha, \mathrm{mc}}$ and $\sHighE(\vx_{n+1}, y) + \epsilon_\mathrm{hoef}$, where  the quantile $q_{\alpha, \mathrm{mc}}$ is computed on the clean scores estimated with MC-sampling without correction. Instead, $\epsilon_\mathrm{hoef}$ is added to account for the difference between the unseen clean test MC-score (exchangeable with the MC-calibration scores) and the upper bound which only bounds the true (non-MC) mean. See \autoref{sec:monte-carlo} for details.
In our case, we compare the corrected quantile $\underline{q}_{\alpha+}$ and the corrected estimate of the input test score $\hat{s}_+(\vx_{n+1}, y)$. Instead of a Hoeffding bound we can use the tighter Bernstein bound for $\hat{s}_+(\vx_{n+1}, y)$ since we have access to it. In addition, to compute $\underline{q}_{\alpha+}$ we use DKW-corrected scores which introduce less error compared to the Hoeffding bound.

\parbold{Feature Poisoning} We find the lower bound quantile $\underline{q_\alpha}$ (\autoref{eq:poison-problem}) using smooth scores (see \autoref{sec:technical-poisoning}, \autoref{eq:poison-problem-mlp}). To apply sample correction, again with an error budget of $\eta$ we divide this budget equally between calibration points. For each calibration point, the CDF bound with correction finds a probabilistic lower bound on the clean smooth score. Since the test scores are computed with MC-estimation, we directly bound the MC-estimated clean calibration scores for exchangeability. Since we do not have access to the clean calibration scores, following \citet{anonymous2023provably} we use Hoeffding's inequality. The corrected quantile is lower than the clean quantile for MC-estimated calibration scores.

\begin{proposition}
    \label{thrm:poisoning-corrected}
    Let $\underline{s} _\mathrm{cdf+}(\tilde{\boldsymbol{x}} _i, y _i) \le \underline{s} _\mathrm{cdf}(\tilde{\boldsymbol{x}} _i, y _i)$ hold with $1 - \eta / (2|\mathcal{D} _\mathrm{cal}|)$ probability for all $(\tilde{\boldsymbol{x}}_i, y_i) \in \mathcal{D} _\mathrm{cal}$.     
    Let $\underline{q}_{\alpha+}$ be the solution to \autoref{eq:poison-problem-mlp} (\autoref{eq:poison-problem} with CDF bounds) for $\alpha = \alpha' - \eta$ with $\underline{s}_i =\underline{s} _\mathrm{cdf +}(\tilde{\boldsymbol{x}} _i, y _i) - \epsilon _\mathrm{hoef}$. Then for each new test point $\boldsymbol{x}_{n+1}$ exchangeable with $\Dcal$ the prediction set defined as $\overline{ \mathcal{C} } (\boldsymbol{x}_{n+1}) = \{y: s _\mathrm{mc}(\boldsymbol{x}_{n+1}, y) \ge \underline{q}_{\alpha+}\}$ has $1 - \alpha'$ coverage. 
\end{proposition}
Here $\epsilon_\mathrm{hoef} = \sqrt{ \log (2/ \eta)/2 | \mathcal{D} _\mathrm{cal} |}$ comes from the Hoeffding inequality. In label poisoning we do not use randomly smoothed scores, therefore sample correction is not needed.

\section{Experiments}
\label{sec:experiments}


For evasion, we compare \method with RSCP \cite{gendler2021adversarially}. Even though the original RSCP is not able to handle sparse or discrete data, we extend it and use it as an additional baseline (see \autoref{sec:rand-smooth-bin}). There are no baselines for poisoning.
Since both RSCP and \method have the same guaranteed coverage we focus on two main metrics: the average size of prediction sets (or efficiency) and the empirical coverage. Ideally, we want the coverage to be concentrated around the nominal $1 - \alpha$. Higher coverage costs larger prediction sets.
In \autoref{sec:additional-experiments} we report additional experiments including the singleton hits ratio metric.
We also consider the maximum perturbation radius such that robust CP has the same set size as standard CP (averaged across test points). This size-preserving $r$ is the largest certified radius which we can get ``for free''.
On all metrics \method outperforms RSCP.

\begin{figure*}[ht]
    \centering
    \input{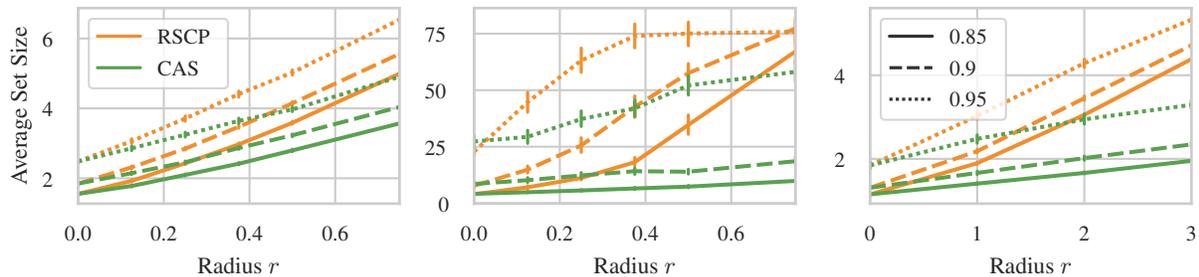}
    \caption{Average set size of \method and RSCP under evasion for (from left to right) \texttt{CIFAR-10}, \texttt{ImageNet} (with TPS), and \texttt{Cora-ML}.}
    \label{fig:all-evasion}
\end{figure*}

\begin{figure*}[t]
    \centering
    \input{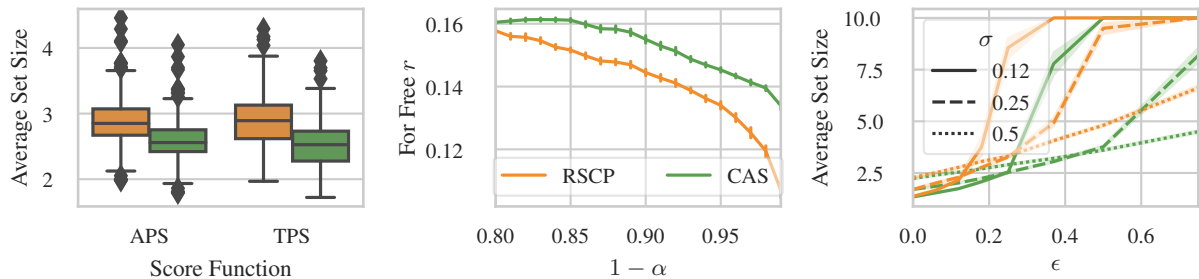}
    \caption{[Left] Set size for $r=0.12$ with different scores. [Middle] Maximum set size-preserving radius (average over test points). Both results are on \texttt{CIFAR-10} dataset and $\sigma=0.25$. [Right] The effect of smoothing parameter $\sigma$ on the set size across a range of radii for \texttt{CIFAR-10} dataset with error correction for $10^4$ samples.}
    \label{fig:score-free-sigma-ablation}
\end{figure*}

\parbold{Setup} We evaluate our method on two image datasets: \texttt{CIFAR-10} \citep{Krizhevsky2009LearningML} and \texttt{ImageNet} \citep{Deng2009ImageNetAL}, and one node-classification (graph) dataset \texttt{Cora-ML} \cite{McCallum2004AutomatingTC}. We used \texttt{ResNet-110} and \texttt{ResNet-50} pretrained on \texttt{CIFAR-10} and \texttt{ImageNet} with noisy data augmentation from \citet{cohen2019certified}. We trained a \texttt{GCN} model \cite{Kipf2017SemiSupervisedCW} for node classification. All models are trained on data augmented with noise. The GNN is trained with 20 nodes per class with stratified sampling as the training set and similarly sampled validation set. 
The size of the calibration set is between 100 and 150 (sparsely labeled setting).
We use APS as the main score function.

For each dataset, we pick a number of test points at random (900 for \texttt{CIFAR-10}, 400 for \texttt{ImageNet}, and 2480 nodes for \texttt{Cora}).
We estimate the expected smooth scores with $10^4$ Monte-Carlo samples. 
All results are an average of 100 runs with exchangeable calibration sampling (details in \autoref{sec:additional-experiments}). 

\parbold{Evasion Certificate} 
The conservative robust sets are necessarily larger than non-robust sets.
Consequently, on \autoref{fig:attacked-data} (left) we observe a higher empirical coverage on clean data compared to the nominal $1 - \alpha$.
The coverage on perturbed inputs which we find with a PGD attack \citep{madry2017towards} is above $1-\alpha$ verifying our theory.
In \autoref{fig:attacked-data} (right) we see that the empirical coverage increases with the certified radius $r$ and is $1 - \alpha$ for $r=0$. 
\method is less needlessly conservative (grows slower with $r$) than RSCP while still providing the same guarantee. 
This leads to improved efficiency (smaller sets) as shown in \autoref{fig:attacked-data} (middle). The set size is slightly higher for perturbed inputs. 

In \autoref{fig:all-evasion} we see that \method's results in smaller prediction sets, across all radii, and all nominal $1-\alpha$ values, and as in \autoref{fig:score-free-sigma-ablation} (left) all scores. The improvement is substantial and also grows with $r$ -- for larger radii it is doubled or even tripled, especially on \texttt{ImageNet} and \texttt{Cora-ML}. 
Similarly, \autoref{fig:score-free-sigma-ablation} (middle) shows that with \method we can consistently certify a larger maximum radius "for free".




\begin{figure*}[h!]
    \centering
    \input{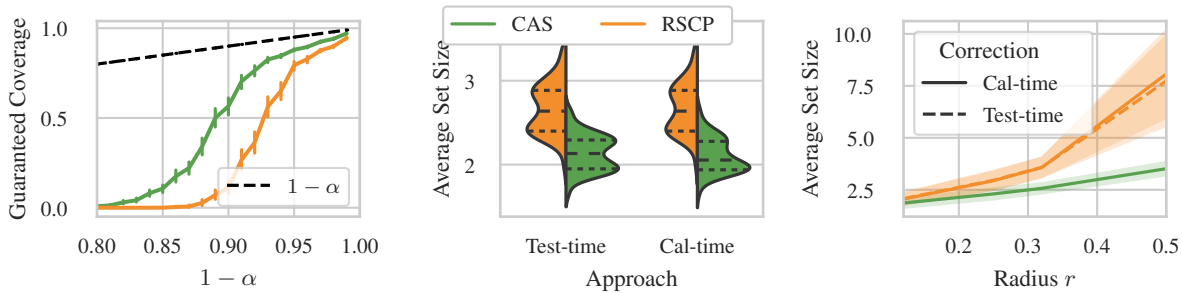}
    \caption{
    [Left] Lower bound $1-\beta$ on the robust coverage of vanilla CP (\autoref{thrm:worst-case-coverage}). 
    \method certifies a larger lower bound.
    [Middle] Distribution of prediction set sizes using the slower test-time vs. the faster calibration-time evasion certificate. 
    [Right] Set sizes for RSCP and \method with account for finite sample error. All results are for \texttt{CIFAR-10} with $\sigma=0.25$.
    }
    \label{fig:three-plots-evasion}
\end{figure*}

\begin{figure*}
    \centering
    \input{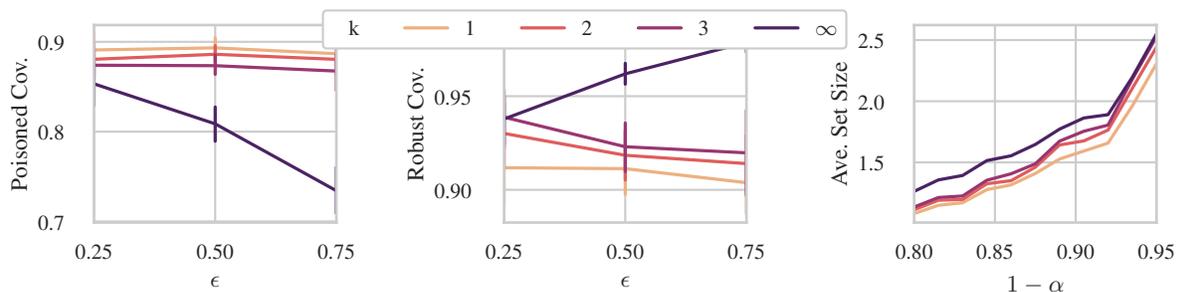}
    \caption{[Left] The coverage of vanilla CP under calibration set with poisoned features. [Middle] The result of robust CP on the same calibration set. [Right] The average set size of the CP robust to feature poisoning for a range of nominal coverages given the clean calibration data, and $r=0.12$. All results are for \texttt{CIFAR-10} dataset with $\sigma=0.25$.}
    \label{fig:poisoning-result}
\end{figure*}

\parbold{Calibration-time evasion}
Following \autoref{thrm:worst-case-coverage} if we use vanilla (non-robust) CP, in the adversarial setup we can certify a lower bound $1-\beta$ on the worst-case robust coverage. In \autoref{fig:three-plots-evasion} (left) we see that the certificate based on \method leads to a better (higher) lower bound. At the same time, \autoref{thrm:worst-case-coverage} implies that we can avoid computing upper bounds for the test points and instead account for the effect of the adversary by choosing a conservative conformal threshold ($\underline{q_\alpha}$) via the lower bound on the calibration scores.
\autoref{fig:three-plots-evasion} (middle) show the set size distribution for test-time vs. calibration-time evasion. The results for RSCP are comparable. \method shows smaller sets for the calibration-time certificate. This approach is also computationally faster especially for datasets with a high number of classes, which is discussed in \autoref{sec:calibration-robustness}.

\parbold{Ablation study}
In \autoref{fig:score-free-sigma-ablation} (right) we study the effect of the smoothing strength as controlled by $\sigma$ in $\gN(\vzero, \sigma^2 \mI)$. 
For all $\sigma$ values and all radii $r$ we get the same $1-\alpha$ coverage guarantee, 
however, there is a clear trade-off for choosing $\sigma$. A smaller amount of smoothing results in a smaller set size in the beginning, but the set sizes grows rapidly by increasing the certified radius. 
In all cases, \method is better than RSCP. Here for each $\sigma$ we use the model that is pretrained on the same noise augmentation.

\parbold{Finite sample correction}
The previous results were without error correction since RSCP did not account for finite-sample errors when estimating the smooth scores with Monte-Carlo samples.
The sets are still asymptotically valid without correction, as confirmed by \autoref{fig:attacked-data} (left); however, correction is necessary for a valid certificate as argued by \citet{anonymous2023provably}.  In \autoref{fig:three-plots-evasion} (right) we see that the size for RSCP quickly explodes, reaching almost all classes ($|\gY|=10$) for large radii, while \method maintains low average size. Moreover, \method has smaller standard deviation across test inputs. \method uses calibration-time correction (see \autoref{sec:monte-carlo}).\looseness=-1

\parbold{Label poisoning}
Next, we study label poisoning where now the attacker can perturb the ground-truth labels of the calibration points. In \autoref{tab:label-poisoning} we see that increasing the budget $k$ leads to predictably larger set size and larger empirical coverage. The difference to the clean calibration set ($k=0$) is minor, showing that provable label robustness comes almost for free for small $k$. 
%
While \citet{einbinder2022ConformalPI} show that standard CP is already naturally robust to random (non-worst case) label noise,
\autoref{tab:label-poisoning} shows that adversarial label noise can break the guarantee even for small budget $k$.
%
\begin{table}[h]
\centering

    \caption{Label poisoning for \texttt{CIFAR-10}.} 
    \vskip 0.5em
    \begin{tabular}{S[table-format=1.0] S[table-format=1.2] S[table-format=1.2] S[table-format=1.2] S[table-format=1.2]} 
        \toprule
              & {Cov.}  & {Vanilla Cov.} & {Robust Cov.} & {Set Size}\\
        {$k$} & {(Clean)} & {(Pert)} & {(Pert)} &  {(Clean)}\\
        \midrule
        0 & 0.897  & 0.897 & 0.897 & 1.41\\
        1 & 0.916  & 0.872 & 0.900 & 1.58\\
        2 & 0.923 & 0.859 & 0.901  & 1.62\\
        \bottomrule
    \end{tabular}
    \label{tab:label-poisoning}
\end{table}

\parbold{Feature poisoning} 
Since there are no baselines that provide robust coverage guarantees under poisoning we can only study the behaviour of \method. First, we consider feature poisoning where the attacker is allowed to change $k$ calibration points which we refer to as the budget, each of which can be perturbed in a given ball $\gB_r(\vx)$ (see \autoref{eq:poison-problem}). 
In \autoref{fig:poisoning-result} (left) we show that the coverage can slightly decrease via poisoning the features with a limited budget. This drop becomes significant when the adversary can perturb all the calibration points.
To poison the data, we run the PGD attack on all calibration points and decide which point to perturb by solving \autoref{eq:poison-problem} (specifically \autoref{eq:poison-problem-mlp}) with maximization goal.
\autoref{fig:poisoning-result} (middle) shows the robustness of \method even under an infinite budget which verifies \autoref{thrm:poisoning}. We also show the set size of robust CP in \autoref{fig:poisoning-result} (right).
We see that as expected a smaller budget $k$ leads to less conservative sets which translates to smaller set sizes. Interestingly, for small $r$ (e.g. $r=0.12$) even with an infinite budget 
the set size does not increase drastically. Making CP robust to poisoning comes at only a small cost. Note that for $k=\infty$ setting each calibration score to its lower bound is one solution to \autoref{eq:poison-problem}, which equals calibration-time evasion. 

Similar to evasion, in \autoref{tab:feature-corrected} we show the results of \method robust to feature poisoning with and without sample correction. For sample correction, we use \autoref{thrm:poisoning-corrected}. Note that label-poisoning does not require sample correction.

\begin{table}
\centering
\begin{tblr}{
  column{3} = {c},
  column{4} = {c},
  column{5} = {c},
  column{6} = {c},
  cell{1}{3} = {c=2}{},
  cell{1}{5} = {c=2}{},
  hline{1,7} = {-}{0.08em},
  hline{3} = {-}{0.05em},
}
\textbf{ } & \textbf{ }     & \textbf{ Emp. Coverage} & & \textbf{Ave. Set Size} &                    \\
$r$    & $k$ & With & Without & With & Without \\
0.12 &  3 & 94.6 & 94.5  & 1.84  & 1.83\\
    & $|\Dcal|$ & 97.7 & 96.3 & 3.17  & 2.47   \\
0.25       &   3              & 94.0                & 94.0               & 1.756 & 1.752\\
           &  $|\Dcal|$            & 99.6                & 98.7               & 7.32                        & 4.48                

\end{tblr}
\caption{CAS for feature poisoning with and without finite-sample correction.}
\label{tab:feature-corrected}
\end{table}





\section{Related Work}
\citet{pmlr-v216-ghosh23a} introduce the notion of probabilistically robust CP. 
Intuitively, their guarantee is w.r.t. the average adversarial input, while for RSCP and our method the guarantee is w.r.t. the worst-case input.
They produce more efficient sets via a quantile of quantiles method -- one quantile considers the adversarial examples around a datapoint and the other finds the CP threshold over the first set of quantiles. This enables a tuneable trade-off between nominal performance and
robustness. 
Our method is orthogonal since we consider exact coverage, and \citet{pmlr-v216-ghosh23a}'s probabilistic robustness can be applied on top of ours.

\citet{cauchois2020robust} propose an approach which returns prediction sets that are robust to distribution shift between the calibration and the test distribution. As input, their method needs an upper bound $\rho$ on the $f$-divergence between the two distributions, which they estimate from data.
In principle, for a given radius $r$ one can derive a suitable $\rho$, however, the resulting sets can be needlessly too conservative.
We can conclude this from the fact that the optimization problem with the resulting $f$-divergence constraint is a relaxation as shown by \citet{Dvijotham2020AFF} in a different context (classification certificates).
\citet{gendler2021adversarially} extensively discuss the differences between RSCP and \citet{cauchois2020robust}'s approach across various settings (e.g. model trained with and without noise) and report better or equal efficiency. With \method outperforming RSCP, we draw similar conclusions by transitivity.
Further discussion is in \autoref{sec:comparison-other-robust}.\looseness=-1

Two concurrent works use the same bound as RSCP, but improve the sets by modifying other aspects of the algorithm. 
\citet{anonymous2023provably} adopt robust conformal training \cite{stutz2022learning} and propose to transform the smooth score (ranking + sigmoid scaling) using an additional holdout set.
\citet{anonymous2024colep} integrate a reasoning component via probabilistic circuits. Both are completely orthogonal to our method and can be directly improved with our CDF bounds.

\citet{Angelopoulos2022ConformalRC} extend conformal prediction to control the expected value of any monotone loss function, including adversarial risk (see Proposition 7). However, they do not propose an algorithm to compute the worst-case adversarial loss.
\citet{einbinder2022ConformalPI} show that standard CP is already robust to \emph{random} label noise, e.g. resulting from wrong annotation or any other natural source of noise. Unlike our work, they do not study robustness to adversarial (worst-case) label perturbations.

\section{Conclusion}

We provide certified robustness for conformal prediction both for evasion and poisoning attacks. We propose a CDF-aware bound on the conformity scores under adversarial perturbation. Our bound is empirically tighter and leads to consistent improvements compared to previous certificates.
We further propose novel certificates against feature and/or label poisoning of the calibration set.
We generalize both results to discrete and binary (sparse) data.
Finally, we show how we can correct for finite-sample error. Our calibration-time approach for robustness to evasion that reduces the inflation of set sizes when correcting for finite samples.
Overall, our method \method yields provably robust yet efficient (small) prediction sets.

\parbold{Limitations}
We identify three main limitations.
First, the coverage guarantee is marginal, which means that it holds on average across the entire input domain. Conditional coverage $\Pr[y \in \gC(\vx) \mid \vx]$ is impossible to achieve without strong assumptions \citep{Barber2019TheLO}. Achieving near-conditional coverage is still and open problem.
This means that CP can have over-coverage or under-coverage for different groups which can be unfair. This holds true for both vanilla CP and robust CP. \citet{lu2022fair} consider a group-conformal variant to equalize coverage across groups, however, unfairness can still be reflected in the set-size. Studying the intersection of robustness and fairness is an exciting future direction.
Second, while randomized smoothing is a powerful and flexible method, estimating empirical statistics requires a large number of Monte-Carlo samples. This can be computationally expensive.
Finally, we assumed that the goal of the attacker is to reduce 
the empirical coverage and designed our certificate to prevent this. However, the attacker may have other goals, e.g. to increase the set size, or to attack only a subset of labels.

\section*{Acknowledgements}
We thank Giuliana Thomanek for feedbacks on our draft.\looseness=-1

\section*{Impact Statement}
This paper presents work whose goal is to advance the field of Machine Learning. There are many potential societal consequences of our work, none of which we feel must be specifically highlighted here.

\nocite{langley00}

\bibliography{paper}

\begin{thebibliography}{41}
\providecommand{\natexlab}[1]{#1}
\providecommand{\url}[1]{\texttt{#1}}
\expandafter\ifx\csname urlstyle\endcsname\relax
  \providecommand{\doi}[1]{doi: #1}\else
  \providecommand{\doi}{doi: \begingroup \urlstyle{rm}\Url}\fi

\bibitem[Abdar et~al.(2021)Abdar, Pourpanah, Hussain, Rezazadegan, Liu, Ghavamzadeh, Fieguth, Cao, Khosravi, Acharya, et~al.]{abdar2021review}
Abdar, M., Pourpanah, F., Hussain, S., Rezazadegan, D., Liu, L., Ghavamzadeh, M., Fieguth, P., Cao, X., Khosravi, A., Acharya, U.~R., et~al.
\newblock A review of uncertainty quantification in deep learning: Techniques, applications and challenges.
\newblock \emph{Information fusion}, 76:\penalty0 243--297, 2021.

\bibitem[Angelopoulos \& Bates(2021)Angelopoulos and Bates]{angelopoulos2021gentle}
Angelopoulos, A.~N. and Bates, S.
\newblock A gentle introduction to conformal prediction and distribution-free uncertainty quantification.
\newblock \emph{arXiv preprint arXiv:2107.07511}, 2021.

\bibitem[Angelopoulos et~al.(2022)Angelopoulos, Bates, Fisch, Lei, and Schuster]{Angelopoulos2022ConformalRC}
Angelopoulos, A.~N., Bates, S., Fisch, A., Lei, L., and Schuster, T.
\newblock Conformal risk control.
\newblock \emph{ArXiv}, abs/2208.02814, 2022.
\newblock URL \url{https://api.semanticscholar.org/CorpusID:251320513}.

\bibitem[Angelopoulos et~al.(2023)Angelopoulos, Bates, et~al.]{angelopoulos2023conformal}
Angelopoulos, A.~N., Bates, S., et~al.
\newblock Conformal prediction: A gentle introduction.
\newblock \emph{Foundations and Trends{\textregistered} in Machine Learning}, 16\penalty0 (4):\penalty0 494--591, 2023.

\bibitem[Athalye et~al.(2018)Athalye, Carlini, and Wagner]{athalye2018obfuscated}
Athalye, A., Carlini, N., and Wagner, D.
\newblock Obfuscated gradients give a false sense of security: Circumventing defenses to adversarial examples.
\newblock In \emph{International conference on machine learning}, pp.\  274--283. PMLR, 2018.

\bibitem[Barber et~al.(2019)Barber, Cand{\`e}s, Ramdas, and Tibshirani]{Barber2019TheLO}
Barber, R.~F., Cand{\`e}s, E.~J., Ramdas, A., and Tibshirani, R.~J.
\newblock The limits of distribution-free conditional predictive inference.
\newblock \emph{Information and Inference: A Journal of the IMA}, 2019.

\bibitem[Bojchevski et~al.(2020)Bojchevski, Gasteiger, and G{\"u}nnemann]{bojchevski2020efficient}
Bojchevski, A., Gasteiger, J., and G{\"u}nnemann, S.
\newblock Efficient robustness certificates for discrete data: Sparsity-aware randomized smoothing for graphs, images and more.
\newblock In \emph{International Conference on Machine Learning}, pp.\  1003--1013. PMLR, 2020.

\bibitem[Cauchois et~al.(2020)Cauchois, Gupta, Ali, and Duchi]{cauchois2020robust}
Cauchois, M., Gupta, S., Ali, A., and Duchi, J.~C.
\newblock Robust validation: Confident predictions even when distributions shift.
\newblock \emph{arXiv preprint arXiv:2008.04267}, 2020.

\bibitem[Cohen et~al.(2019)Cohen, Rosenfeld, and Kolter]{cohen2019certified}
Cohen, J., Rosenfeld, E., and Kolter, Z.
\newblock Certified adversarial robustness via randomized smoothing.
\newblock In \emph{international conference on machine learning}, pp.\  1310--1320. PMLR, 2019.

\bibitem[Deng et~al.(2009)Deng, Dong, Socher, Li, Li, and Fei-Fei]{Deng2009ImageNetAL}
Deng, J., Dong, W., Socher, R., Li, L.-J., Li, K., and Fei-Fei, L.
\newblock Imagenet: A large-scale hierarchical image database.
\newblock \emph{2009 IEEE Conference on Computer Vision and Pattern Recognition}, pp.\  248--255, 2009.
\newblock URL \url{https://api.semanticscholar.org/CorpusID:57246310}.

\bibitem[Dvijotham et~al.(2020)Dvijotham, Hayes, Balle, Kolter, Qin, Gy{\"o}rgy, Xiao, Gowal, and Kohli]{Dvijotham2020AFF}
Dvijotham, K., Hayes, J., Balle, B., Kolter, Z., Qin, C., Gy{\"o}rgy, A., Xiao, K.~Y., Gowal, S., and Kohli, P.
\newblock A framework for robustness certification of smoothed classifiers using f-divergences.
\newblock In \emph{International Conference on Learning Representations}, 2020.
\newblock URL \url{https://api.semanticscholar.org/CorpusID:213452491}.

\bibitem[Dvoretzky et~al.(1956)Dvoretzky, Kiefer, and Wolfowitz]{dvoretzky1956asymptotic}
Dvoretzky, A., Kiefer, J., and Wolfowitz, J.
\newblock Asymptotic minimax character of the sample distribution function and of the classical multinomial estimator.
\newblock \emph{The Annals of Mathematical Statistics}, pp.\  642--669, 1956.

\bibitem[Einbinder et~al.(2022)Einbinder, Bates, Angelopoulos, Gendler, and Romano]{einbinder2022ConformalPI}
Einbinder, B.-S., Bates, S., Angelopoulos, A.~N., Gendler, A., and Romano, Y.
\newblock Conformal prediction is robust to label noise.
\newblock \emph{ArXiv}, abs/2209.14295, 2022.
\newblock URL \url{https://api.semanticscholar.org/CorpusID:262091979}.

\bibitem[Fey \& Lenssen(2019)Fey and Lenssen]{Fey/Lenssen/2019}
Fey, M. and Lenssen, J.~E.
\newblock Fast graph representation learning with {PyTorch Geometric}.
\newblock In \emph{ICLR Workshop on Representation Learning on Graphs and Manifolds}, 2019.

\bibitem[Gendler et~al.(2021)Gendler, Weng, Daniel, and Romano]{gendler2021adversarially}
Gendler, A., Weng, T.-W., Daniel, L., and Romano, Y.
\newblock Adversarially robust conformal prediction.
\newblock In \emph{International Conference on Learning Representations}, 2021.

\bibitem[Ghosh et~al.(2023)Ghosh, Shi, Belkhouja, Yan, Doppa, and Jones]{pmlr-v216-ghosh23a}
Ghosh, S., Shi, Y., Belkhouja, T., Yan, Y., Doppa, J., and Jones, B.
\newblock Probabilistically robust conformal prediction.
\newblock In Evans, R.~J. and Shpitser, I. (eds.), \emph{Proceedings of the Thirty-Ninth Conference on Uncertainty in Artificial Intelligence}, volume 216 of \emph{Proceedings of Machine Learning Research}, pp.\  681--690. PMLR, 31 Jul--04 Aug 2023.
\newblock URL \url{https://proceedings.mlr.press/v216/ghosh23a.html}.

\bibitem[Guo et~al.(2017)Guo, Pleiss, Sun, and Weinberger]{guo2017calibration}
Guo, C., Pleiss, G., Sun, Y., and Weinberger, K.~Q.
\newblock On calibration of modern neural networks.
\newblock In \emph{International conference on machine learning}, pp.\  1321--1330. PMLR, 2017.

\bibitem[Huang et~al.(2023)Huang, Jin, Candes, and Leskovec]{huang2023uncertainty}
Huang, K., Jin, Y., Candes, E., and Leskovec, J.
\newblock Uncertainty quantification over graph with conformalized graph neural networks.
\newblock \emph{arXiv preprint arXiv:2305.14535}, 2023.

\bibitem[Kang et~al.(2024)Kang, G{\"u}rel, Li, and Li]{anonymous2024colep}
Kang, M., G{\"u}rel, N.~M., Li, L., and Li, B.
\newblock {COLEP}: Certifiably robust learning-reasoning conformal prediction via probabilistic circuits.
\newblock In \emph{The Twelfth International Conference on Learning Representations}, 2024.
\newblock URL \url{https://openreview.net/forum?id=XN6ZPINdSg}.

\bibitem[Kipf \& Welling(2017)Kipf and Welling]{Kipf2017SemiSupervisedCW}
Kipf, T. and Welling, M.
\newblock Semi-supervised classification with graph convolutional networks.
\newblock \emph{ArXiv}, abs/1609.02907, 2017.

\bibitem[Krizhevsky(2009)]{Krizhevsky2009LearningML}
Krizhevsky, A.
\newblock Learning multiple layers of features from tiny images.
\newblock 2009.
\newblock URL \url{https://api.semanticscholar.org/CorpusID:18268744}.

\bibitem[Kumar et~al.(2020)Kumar, Levine, Feizi, and Goldstein]{kumar2020certifying}
Kumar, A., Levine, A., Feizi, S., and Goldstein, T.
\newblock Certifying confidence via randomized smoothing.
\newblock \emph{Advances in Neural Information Processing Systems}, 33:\penalty0 5165--5177, 2020.

\bibitem[Lecuyer et~al.(2019)Lecuyer, Atlidakis, Geambasu, Hsu, and Jana]{lecuyer2019certified}
Lecuyer, M., Atlidakis, V., Geambasu, R., Hsu, D., and Jana, S.
\newblock Certified robustness to adversarial examples with differential privacy.
\newblock In \emph{2019 IEEE symposium on security and privacy (SP)}, pp.\  656--672. IEEE, 2019.

\bibitem[Lee et~al.(2019)Lee, Yuan, Chang, and Jaakkola]{lee2019tight}
Lee, G.-H., Yuan, Y., Chang, S., and Jaakkola, T.
\newblock Tight certificates of adversarial robustness for randomly smoothed classifiers.
\newblock \emph{Advances in Neural Information Processing Systems}, 32, 2019.

\bibitem[Levine \& Feizi(2021)Levine and Feizi]{levine2021improved}
Levine, A. and Feizi, S.
\newblock Improved, deterministic smoothing for $l_1$ certified robustness.
\newblock 2021.

\bibitem[Li et~al.(2023)Li, Xie, and Li]{li2023sok}
Li, L., Xie, T., and Li, B.
\newblock Sok: Certified robustness for deep neural networks.
\newblock 2023.

\bibitem[Lu et~al.(2022)Lu, Lemay, Chang, H{\"o}bel, and Kalpathy-Cramer]{lu2022fair}
Lu, C., Lemay, A., Chang, K., H{\"o}bel, K., and Kalpathy-Cramer, J.
\newblock Fair conformal predictors for applications in medical imaging.
\newblock In \emph{Proceedings of the AAAI Conference on Artificial Intelligence}, volume~36, pp.\  12008--12016, 2022.

\bibitem[Madry et~al.(2017)Madry, Makelov, Schmidt, Tsipras, and Vladu]{madry2017towards}
Madry, A., Makelov, A., Schmidt, L., Tsipras, D., and Vladu, A.
\newblock Towards deep learning models resistant to adversarial attacks.
\newblock \emph{arXiv preprint arXiv:1706.06083}, 2017.

\bibitem[McCallum et~al.(2004)McCallum, Nigam, Rennie, and Seymore]{McCallum2004AutomatingTC}
McCallum, A., Nigam, K., Rennie, J. D.~M., and Seymore, K.
\newblock Automating the construction of internet portals with machine learning.
\newblock \emph{Information Retrieval}, 3:\penalty0 127--163, 2004.

\bibitem[Mujkanovic et~al.(2022)Mujkanovic, Geisler, G{\"u}nnemann, and Bojchevski]{mujkanovic2022defenses}
Mujkanovic, F., Geisler, S., G{\"u}nnemann, S., and Bojchevski, A.
\newblock Are defenses for graph neural networks robust?
\newblock \emph{Advances in Neural Information Processing Systems}, 35:\penalty0 8954--8968, 2022.

\bibitem[Paszke et~al.(2019)Paszke, Gross, Massa, Lerer, Bradbury, Chanan, Killeen, Lin, Gimelshein, Antiga, Desmaison, Kopf, Yang, DeVito, Raison, Tejani, Chilamkurthy, Steiner, Fang, Bai, and Chintala]{NEURIPS2019_9015}
Paszke, A., Gross, S., Massa, F., Lerer, A., Bradbury, J., Chanan, G., Killeen, T., Lin, Z., Gimelshein, N., Antiga, L., Desmaison, A., Kopf, A., Yang, E., DeVito, Z., Raison, M., Tejani, A., Chilamkurthy, S., Steiner, B., Fang, L., Bai, J., and Chintala, S.
\newblock Pytorch: An imperative style, high-performance deep learning library.
\newblock In \emph{Advances in Neural Information Processing Systems 32}, pp.\  8024--8035. Curran Associates, Inc., 2019.

\bibitem[Romano et~al.(2020)Romano, Sesia, and Cand{\`e}s]{Romano2020ClassificationWV}
Romano, Y., Sesia, M., and Cand{\`e}s, E.~J.
\newblock Classification with valid and adaptive coverage.
\newblock \emph{arXiv: Methodology}, 2020.

\bibitem[Sadinle et~al.(2018)Sadinle, Lei, and Wasserman]{Sadinle2018LeastAS}
Sadinle, M., Lei, J., and Wasserman, L.~A.
\newblock Least ambiguous set-valued classifiers with bounded error levels.
\newblock \emph{Journal of the American Statistical Association}, 114:\penalty0 223 -- 234, 2018.

\bibitem[Salman et~al.(2019)Salman, Li, Razenshteyn, Zhang, Zhang, Bubeck, and Yang]{salman2019provably}
Salman, H., Li, J., Razenshteyn, I., Zhang, P., Zhang, H., Bubeck, S., and Yang, G.
\newblock Provably robust deep learning via adversarially trained smoothed classifiers.
\newblock \emph{Advances in Neural Information Processing Systems}, 32, 2019.

\bibitem[Silva \& Najafirad(2020)Silva and Najafirad]{silva2020opportunities}
Silva, S.~H. and Najafirad, P.
\newblock Opportunities and challenges in deep learning adversarial robustness: A survey.
\newblock \emph{arXiv preprint arXiv:2007.00753}, 2020.

\bibitem[Stutz et~al.(2022)Stutz, Dvijotham, Cemgil, and Doucet]{stutz2022learning}
Stutz, D., Dvijotham, K.~D., Cemgil, A.~T., and Doucet, A.
\newblock Learning optimal conformal classifiers.
\newblock In \emph{International Conference on Learning Representations}, 2022.
\newblock URL \url{https://openreview.net/forum?id=t8O-4LKFVx}.

\bibitem[Teng et~al.(2023)Teng, Wen, Zhang, Bengio, Gao, and Yuan]{teng2023predictive}
Teng, J., Wen, C., Zhang, D., Bengio, Y., Gao, Y., and Yuan, Y.
\newblock Predictive inference with feature conformal prediction.
\newblock In \emph{The Eleventh International Conference on Learning Representations}, 2023.
\newblock URL \url{https://openreview.net/forum?id=0uRm1YmFTu}.

\bibitem[Vovk et~al.(2005)Vovk, Gammerman, and Shafer]{Vovk2005AlgorithmicLI}
Vovk, V., Gammerman, A., and Shafer, G.
\newblock Algorithmic learning in a random world.
\newblock 2005.

\bibitem[Yan et~al.(2024)Yan, Romano, and Weng]{anonymous2023provably}
Yan, G., Romano, Y., and Weng, T.-W.
\newblock Provably robust conformal prediction with improved efficiency.
\newblock In \emph{The Twelfth International Conference on Learning Representations}, 2024.
\newblock URL \url{https://openreview.net/forum?id=BWAhEjXjeG}.

\bibitem[Yuan et~al.(2019)Yuan, He, Zhu, and Li]{yuan2019adversarial}
Yuan, X., He, P., Zhu, Q., and Li, X.
\newblock Adversarial examples: Attacks and defenses for deep learning.
\newblock \emph{IEEE transactions on neural networks and learning systems}, 30\penalty0 (9):\penalty0 2805--2824, 2019.

\bibitem[Zargarbashi et~al.(2023)Zargarbashi, Antonelli, and Bojchevski]{zargarbashi2023GraphCP}
Zargarbashi, S.~H., Antonelli, S., and Bojchevski, A.
\newblock Conformal prediction sets for graph neural networks.
\newblock In \emph{Proceedings of the 40th International Conference on Machine Learning}, 2023.
\newblock URL \url{https://openreview.net/forum?id=zGf8J0bNfX}.

\end{thebibliography}
\bibliographystyle{icml2024}

\newpage
\appendix
\onecolumn

\section{More On Conformal Prediction}
\label{sec:conformal-more}

\parbold{Conformity vs. non-conformity scores} 
As mentioned in \autoref{sec:background}, for CP we need to define a score function that quantifies the agreement between the input and each label. Equivalently, one can define CP with a \emph{non}-conformity score function that captures disagreement instead. In this case, the conformal threshold is the $1 - \alpha$ quantile of the calibration true scores. Similarly, in the test time, labels with score \emph{less} than the threshold are included in the prediction set. Both approaches are equivalent up to a change in the sign of the scores.  
The latter setup is used in \citep{gendler2021adversarially} and is equivalent to our implementation that uses conformity scores. Our choice of agreement score is due to simplicity.

\parbold{Score function}
In \autoref{sec:background} we mentioned that conformal prediction returns guaranteed sets regardless of the score function employed. Specifically, any score function maintaining the exchangeability (between calibration and test) is viable. 
In brief, the exchangeability of random variable $Z_1, \dots, Z_n$ means that the joint distribution of the variables is insensitive to the order/index. In other words for any permutation function $\psi:[n]\mapsto[n]$ we have $\prob{Z_1, \dots, Z_n} = \prob{Z_{\psi(1)}, \dots, Z_{\psi(n)}}$. Assuming the calibration set to be exchangeably sampled from the data distribution, any permutation equivariant transformation on the data still preserves the exchangeability. Conclusively, the smooth scores from \citet{gendler2021adversarially} and  \citet{bojchevski2020efficient} are both permutation equivariant (the smoothing applies similarly to all calibration and test points regardless of their order). Therefore, smoothing scores maintains exchangeability.

While any score function preserving the exchangeability maintains the conformal guarantee, better scores result in better performance with respect to the metric of interest. For instance, even a function that returns uniform conformity scores at random provides a valid guarantee, although the prediction sets will be large. 

%
Various score functions are proposed in the literature of conformal classification ranging from simple softmax function on top of model's result \cite{Sadinle2018LeastAS}, to more complex functions leveraging information from embedding spaces of the model \cite{teng2023predictive}, or from the confidence of adjacent datapoints within a network structure \citep{zargarbashi2023GraphCP}. The expected score within the smoothing scheme around an input is no exception as it only involves the datapoint itself and applies symmetrically to all datapoints. Similar conditions hold for any approximation of that expectation e.g. the mean of Monte-Carlo samples. See \S B in \citet{anonymous2023provably} for a longer discussion.


\parbold{Effect of the calibration set size}
With a calibration set exchangeably sampled from the data distribution (infinite samples), conformal prediction provides a marginal coverage of at least $1 - \alpha$ (\autoref{eq:cp-guarantee}).  This probability is also upper bounded by $1 - \alpha + 1/(n+1)$. Precisely, the coverage is distributed as $\mathrm{Beta}(n+1 - l, l)$ with $l = \mathfloor{(n+1)\alpha}$.

For a finite set of points and an exchangeably sampled calibration subset, e.g. transductive node-classification, \citet{huang2023uncertainty} show that the coverage probability, $\mathrm{Cov}(\gD) = (1/|\gD|)\sum_{(\vx_i, y_i) \in \gD}\1[y_i \in \gC(\vx_i)]$ is distributed as \begin{align}
    \prob{\mathrm{Cov}(\gD) \le t} = 1 - \Phi_{\mathrm{HG}}(\mathfloor{\alpha(n+1)} - 1; M+N, N, \mathceil{Mt} + \mathfloor{\alpha(n+1)})
\end{align} Where $M=|\gD|$, $N=|\Dcal|$ is the size of the calibration set, and $\Phi_{\mathrm{HG}}(P, p, K)$ is the CDF function of hypergeometric distribution of population $P$, sample size $p$, and $K$ successful samples within the population. 


This means that the coverage probability on standard CP is concentrated around $1 - \alpha$. It also means that the variance around $1-\alpha$ decrease as the size of $\Dcal$ increases. When moving the threshold from $q_\alpha$ to any other value $q'$ within the domain of the score function (as in poisoning), the new threshold will correspond to another quantile $\beta = \invquantile{q'}{\Dcal}$ and the coverage will be similarly concentrated around $1 - \beta$. 

Access to a large calibration set (e.g. 1000 points) is unrealistic. Even with a large set of labeled points, there is an open question of whether to use a portion of it for training the model toward better accuracy which can help even in the efficiency of CP. While we ran our experiments with the sparse labeled setting, increasing the size of the calibration set will result in similar values on average but the results will be more concentrated following the distribution of conformal probability.

\parbold{Conservative coverage} 
Both RSCP and \method result in an empirical coverage higher than $1 - \alpha$ for clean data. This is since the vanilla prediction set is a subset of their conservative prediction set. The empirical coverage for RSCP is even higher compared to \method since it uses looser bounds on the score and the prediction sets are \emph{unnecessarily} more conservative. Higher empirical coverage is gained by larger prediction sets; therefore the goal of Robust CP is to find conservative sets that cover the worst-case perturbed input with higher than $1 - \alpha$ probability but not by increasing the set size significantly. 

\parbold{One-sided robust guarantee} Although CP comes with a two-sided coverage guarantee (upper and lower bound on the coverage probability), our robust coverage guarantee is one-sided -- we only guarantee that the coverage is larger than $1 - \alpha$. 
The standard two-sided guarantee relies on exchangeability. However, since the adversary might perturb each point differently, i.e. we have a non-symmetric mapping from clean $\boldsymbol{x}$ to perturbed $\tilde{\boldsymbol{x}}$; therefore, the perturbed points are no longer exchangeable. Another strategy to obtain the second side, would be to compute $\max_{\tilde{\boldsymbol{x}} \in \mathcal{B}(\boldsymbol{x})} \overline{s}(\tilde{\boldsymbol{x}}, y)$ which needs access to the clean test point. Given the difficulty, we leave computing two-sided guarantees for future work.

\subsection{Impelementation Details}
We based our implementation on PyTorch \citep{NEURIPS2019_9015} and Pytorch Geometric \citep{Fey/Lenssen/2019}. We run all our experiments both on CPU (Intel(R) Xeon(R) Platinum 8368 CPU @ 2.40GHz) and, and on GPU (NVIDIA A100-SXM4-40GB).

\section{Faster Evasion-Robustness via Calibration-time Bound}
\label{sec:calibration-robustness}
The evasion-robust CP algorithm (see \autoref{sec:method}) requires an estimation of the expected smooth score for \begin{enumerate*}[label=(\roman*)]
    \item the true class for all calibration points,
    \item and all classes for each test point.
\end{enumerate*}
Moreover, for the standard evasion-aware robustness, we need to additionally compute adversarial upper bounds (solutions to \autoref{eq:certificate-smooth-cdf}) within the threat model for all classes of all test points.
This upper bound has a closed-form for continuous data, and an efficient algorithm for binary/discrete data (see \autoref{sec:rand-smooth-bin}). Nonetheless, it can be beneficial to reduce the overall runtime.
Let $t_{\mathrm{bound}}$ be the time complexity for the upper bound computation for a single $(\vx, y)$ and $t_{\mathrm{MC}}$ be the time complexity of approximating the expected smooth score with $M$ Monte-Carlo samples. With $n$ calibration points and $c$ classes, we need $\gO(n \times t_{\mathrm{MC}})$ time
for calibration, including the quantile computation. Then, for each test point we need $\gO(c \times t_{\mathrm{MC}}\times t_{\mathrm{bound}})$ time.

We define a computationally more efficient and robust alternative built upon \autoref{thrm:worst-case-coverage} in which we offload the computational overhead from the test set to the calibration set. \autoref{thrm:worst-case-coverage} gives a worst-case coverage lower bound for vanilla CP -- even if we evaluate vanilla CP with smooth (but not upper bounded) scores.
Alternatively, we can find a conservative quantile that results in a certified $1 - \alpha$ coverage probability for the worst case input. We call this approach the faster evasion method.


This method of producing prediction sets significantly reduces the computation in two ways: \begin{enumerate*}[label=(\roman*)]
    \item instead of test points (which are larger in number), we compute the upper bounds on calibration points, 
    \item instead of computing the upper bound for all classes, we only compute it for the true class.
\end{enumerate*} 
Thus, we need $\gO(n \times t_{\mathrm{MC}} \times t_\mathrm{bound})$ for calibration, and $\gO(c \times t_{\mathrm{MC}})$ for each test point. In practical scenarios where the test set (during deployment) is larger than the calibration set, the computational savings of the faster approach become significant, especially for tasks with a large number of classes (e.g. \texttt{ImageNet} with 1000 classes).
As shown in \autoref{tab:comparison_runtime} we gain a significant speed up (more than 3X) on \texttt{CIFAR-10} with 204 calibration points and just 100 test points. Here we have gains despite using a relatively tiny test set (even smaller than the calibration set) since we have $c=10$ classes.
Similar, and even better speed-ups can be achieved for datasets with a larger test set and larger number of classes.

\begin{table}[h!]
    \centering
    \caption{Run-time comparison between test-time (slower) and calibration-time (faster) upper bound computation. The result is for \texttt{CIFAR-10} with $10^4$ number of Monte Carlo samples. Here, $m$ is the number of test samples.}
    \vskip 1em
    \label{tab:comparison_runtime} 
    \begin{tabular}{lcccc}
        \toprule
        & \multicolumn{2}{c}{Time (seconds)} & {No. Datapoints} \\
        \cmidrule(lr){2-3}
        {Runtime} & {Standard Evasion Robust Sets} & {Faster Evasion Robust Sets} & \\
        \midrule
        Calibration & 0.15 $\gO(n \times t_{\mathrm{MC}})$  & 0.79 $\gO(n \times t_{\mathrm{MC}} \times t_\mathrm{b})$ & 204 \\ 
        Testing        & 2.93 $\gO(m\times c \times t_{\mathrm{MC}}\times t_\mathrm{b})$ & 0.15 $\gO(m\times c \times t_{\mathrm{MC}})$ & 100 \\ 
        \midrule
        Total           & 3.08  & 0.94 &  \\ 
        \bottomrule
    \end{tabular}
\end{table}

\section{Technical Details On Randomized Smoothing}
\label{sec:rand-smooth-bin}

RSCP uses the closed-form solution to \autoref{eq:certificate-smooth} as an upperbound on the score function within the $L_2$ perturbation radius (details are in \autoref{sec:details-rscp}). The same equation can be used to address other perturbation schemes (e.g. perturbations for sparse data). We use the results from \citet{bojchevski2020efficient} to find extend RSCP to sparse and discrete data and use it as a baseline.

To apply randomized smoothing we need to define a smoothing scheme $\smoothf(\cdot)$ -- a probabilistic function that adds random noise to the input. Given any score function $s$, we define $\hat{s}(\vx, y) = \E[s(\smoothf(\vx), y)]$. Now $\prob{\smoothf(\vx) = \vz}$ is the probability of visiting some $\vz$ in the domain by smoothing from $\vx$. For a continuous data we use Gaussian smoothing where $\smoothf(\vx) = \vx + \delta$ with $\delta \sim \gN(\vzero, \sigma^2\mI)$ coming from an isotropic Gaussian distribution with zero mean and variance $\sigma$. 
We can compute the adversarial upper bounds using the closed-form expressions from \citet{kumar2020certifying} (see e.g. \autoref{eq:gaussian_cdf} in \autoref{sec:efficienct-bounds}).

For binary data, following \citet{bojchevski2020efficient}, we use the following smoothing function: \begin{align}
    \prob{\smoothf(\vx)[i] \neq \vx[i]} = p_{\vx[i]}
\end{align}
This means that $\smoothf$ toggles each 1-bit of $\vx$ with probability $p_1$ and each 0-bit with $p_0$. This distinction allows us to preserve sparsity by specifying a lower $p_0$. Setting $p_1 = p_0 = p$ we have the special case of flipping each bit with the same probability $p$. Similarly \citet{bojchevski2020efficient} generalizes the binary case to the discrete case. Assuming that $\vx \in \gX_K = \mathset{0, 1, \dots, K}^d$ the sparsity aware randomization scheme is defined as \begin{align}
    \prob{\smoothf(\vx)_i = k} = \begin{cases}
        \left( \frac{p_0}{K - 1} \right)^{(\vx[i] \neq k)}(1 - p_0)^{(\vx[i] = k)} & \vx[i] = 0 \\
        \left( \frac{p_1}{K - 1} \right)^{(\vx[i] \neq k)}(1 - p_1)^{(\vx[i] = k)} & \vx[i] \neq 0
    \end{cases}
\end{align} that flips any zero bit with probability $p_0$ and any non-zero bit with $p_1$ to any other $(K - 1)$ possible value.

For the baseline bound we can rewrite \autoref{eq:certificate-smooth} as a linear program by partitioning the input space $\gX$ into regions of constant likelihood ratio \citep{lee2019tight}.
Let $\gX = \bigcup_i \gR_i$ and $\gR_i \bigcap \gR_j = \emptyset$ be a partitioning into disjoint regions of constant likelihood ratio such that for ever $\vz \in  \gR_i$ it holds $\frac{\prob{\smoothf(\vx)=\vz}}{\prob{\smoothf(\pertx)=\vz}}=c_i$ for some constant $c_i$. Let $t_i = \prob{\smoothf(\vx) \in \gR_i}$ and $\tilde{t}_i = \prob{\smoothf(\pertx) \in \gR_i}$ for for each region $\gR_i$. Then \autoref{eq:certificate-smooth} is equivalent to:
\begin{align}
    \label{eq:lp_baseline}
\max_{\vh} \vh^T \tilde{\vt}  \quad\text{ s.t. }\quad \vh^T \vt = p, \quad 0 \leq \vh \le 1
\end{align}
where $\vh \in [0, 1]^I$ is the vector that we are optimizing over corresponding to the score function $h \in \gH$, $\vt$ and $\tilde{\vt}$ are the vectors with $t_i$ and $\tilde{t}_i$ as elements, and $I=r_a+r_d+1$ is the number of regions. Note, that by replacing the constraint with $a \leq \vh \le b$ we can handle score functions that are bounded in $[a, b]$.
The exact solution to this LP can be easily obtained
with a simple algorithm. We visit each region in increasing order w.r.t. $c_i$ where
\begin{align}
    c_i = \left[\frac{p_0}{1-p_1}\right]^{i-r_d} \left[\frac{p_1}{1-p_0}\right]^{i-r_a}
\end{align}
and assign $h_i = 1$ for all regions $\gR_i$ until the budget constraint is met, and $h_i = 0$ for the remaining regions, with the exception of the region in between where $h_i$ is a value between $0$ and $1$ such that the equality constraint is exactly met.
Since the likelihood ratios $c_i$ are monotonic in $i$, the regions are automatically sorted so the solution to the LP can be obtain in linear $O(I)$ time. See \citet{bojchevski2020efficient} for more details and the pseudo-code.

For the CDF-based bound we can similarly rewrite \autoref{eq:certificate-smooth-cdf} as the following linear program:
\begin{align}
\label{eq:lp_cdf}
\max_{\mH} b_m - \mH  \tilde{\vt}  \vd 
\quad\text{ s.t. }\quad \mH \vt = \vp, \quad 0 \leq \mH \le 1
\end{align}
where $\mH \in [0, 1]^{(m-1) \times I}$ is the matrix that we are optimizing over with $H_{ji}$ being the score that we assign to the $j$-th bin and the $i$-th region, $\vd$ is the vector of bin widths such that $d_j = b_{j} - b_{j-1}$, and $\vp$ is a vector where $p_i = \prob{s(\smoothf(\vx), y) \le b_i}$. 
Intuitively, for each bin and each region the worst-case score function $h \in \gH$ assigns the same score to all $\vz$ in that region since the likelihood ratio is constant.
As before we have a simple algorithm to obtain the exact solution to this LP. Observe that \autoref{eq:lp_cdf} can be decomposed into $m-1$ separate LPs similar to \autoref{eq:lp_baseline} which can be solved in parallel using the same algorithm as above. The reason is that there is no interaction between the different bins (different rows of $\mH$) in neither the constraint nor the objective function. Therefore, the solution can be obtain in $O(m \times I)$ with serial computation and $O(I)$ with parallel computation.

\parbold{Tightness} All four bounds are tight, i.e. cannot be improved unless we make additional assumption or provide additional constraints. The reason is that there exists a base score function $s$ such that when relaxing to $h \in \gH$ we get an equality in \autoref{eq:upper_bound_trivial}.
See \citet{kumar2020certifying} for a discussion of why the two Gaussian bounds are tight when certifying the confidence of a classifier and observe that their analysis immediately applies to our score functions. Similarly, the two discrete bounds are tight since there exists an $s$ for which we obtain equality. The $s$ can be constructed using the optimal $\vh^*$ from the problem in \autoref{eq:lp_baseline} and similarly for \autoref{eq:lp_cdf}.

\section{Supplementary to Theoretical Support}

\subsection{Proofs}
\label{sec:proofs}
\begin{proof}[Proof of \autoref{thrm:conservative-guarantee}]
    Given the exchangeability of $(\vx_{n+1}, y_{n+1})$ with the calibration set, \autoref{eq:cp-guarantee} holds for the clean point. 
    Since $\pertx_{n+1} \in \gB(\vx_{n+1})$ we have $\forall y_i: \sHigh(\pertx_{n+1}, y_i) \ge s(\vx_{n+1}, y_i)$. By the definition of CP for any label $y_i$ we have
    \begin{align*}
        y_i \in \gC(\vx_{n+1}) \Rightarrow s(\vx_{n+1}, y_i) \ge q_\alpha \\
        \vx_{n+1} \in \gB(\pertx_{n+1}) \Rightarrow \sHigh(\pertx_{n+1}, y_i) \ge  s(\vx_{n+1}, y_i) \ge q_\alpha \Rightarrow \gC(\vx_{n+1})\subseteq \conservativeSet(\pertx)
    \end{align*}
    Which clearly implies that $\prob{y_{n+1} \in \conservativeSet(\pertx)} \ge \prob{y_{n+1} \in \gC(\vx)} \ge 1-\alpha$.
\end{proof}

\begin{proof}[Proof of \autoref{thrm:poisoning}]
    By definition $\Dcal \in \setgB_{k, \gB}(\Dcal)$; therefore $q_\alpha$ is a feasible solution to \autoref{eq:poison-problem} and we have $\underline{q_\alpha} \leq q_\alpha$.
    It follows that  $\gC_\alpha(\vx) \subseteq \overline{\gC}_\alpha(\vx)$
    where $\gC_\alpha(\vx) = \mathset{y_i: s(\vx, y_i) \ge q_\alpha}$ and $\overline{\gC}_\alpha(\vx) = \mathset{y_i: s(\vx, y_i) \ge \underline{q_\alpha}}$.
    Since $\prob{y_{n+1} \in \gC_\alpha(\vx)} \ge 1-\alpha$
    due to exchangeability it follows that $\prob{y_{n+1} \in \overline{\gC}_\alpha(\vx)} \ge 1-\alpha$.
    In summary the following chain of inequlities hold:
    \begin{align*}
        \forall y_i: y_i \in C(\vx_{n+1}) \Rightarrow s(\vx_{n+1}, y_i) \ge q_\alpha\\
        \underline{q_\alpha} \le q_\alpha \Rightarrow s(\vx_{n+1}, y_i) \ge q_\alpha \ge \underline{q_\alpha} \Rightarrow y_i \in \conservativeSet(\vx_{n+1})
    \end{align*}
    
\end{proof}

\begin{proof}[Proof of \autoref{thrm:worst-case-coverage}]
    Setting $\underline{q_\alpha} = \quantile{\alpha}{\mathset{\sLow(\vx_i, y_i): (\vx_i, y_i) \in \Dcal}}$ we have:
    \begin{align*}
        \prob{\hat{s}(\pertx_{n+1}, y_{n+1}) \ge \underline{q_\alpha}} \ge \prob{\sLow(\vx_{n+1}, y_{n+1}) \ge \underline{q_\alpha}}  & \quad \text{Lower bound within the threat model}\\
        \ge 1 - \alpha & \quad \text{Exchangeability between lower bounds}\\
    \end{align*}
    Alternatively, the vanilla CP is calibrated with quantile $q_\alpha = \quantile{\alpha}{\mathset{\hat{s}(\vx_i, y_i): (\vx_i, y_i) \in \Dcal}}$. The probability of a given potentially perturbed $\pertx_{n+1}$ being covered is:
    \begin{align*}
        \prob{y_{n+1} \in \gC_\alpha(\pertx_{n+1})} = \prob{\hat{s}(\pertx_{n+1}, y_{n+1})\ge q_\alpha} & \quad \text{Definition of CP}\\
        \ge \prob{\sLow(\vx_{n+1}, y_{n+1}) \ge q_\alpha} & \quad \text{Lower bound within the threat model}
    \end{align*}
    Let $\beta = \invquantile{q_\alpha}{\mathset{\sLow(\vx_i, y_i): (\vx_i, y_i) \in \Dcal}}$. If $\sLow$ is computed symmetrically -- indices are invariant to $\sLow$, then $\sLow(\vx_{n+1})$ and $\mathset{\sLow(\vx_i, y_i): (\vx_i, y_i) \in \Dcal}$ are exchangeable. Hence, via quantile lemma we have:
    \begin{align*}
        \prob{\sLow(\vx_{n+1}, y_{n+1}) \ge q_\alpha} \ge 1 - \beta
    \end{align*}
\end{proof}

\begin{proof}[Proof of \autoref{thrm:conformal-mc}]
    Since for each calibration point $\sLowCdfE(\vx_i, y_i) \le \sLowCdf(\vx_i, y_i)$ has at most $\frac{\eta}{2|\Dcal|}$ failure probability following holds with $1 - \eta/2$ probability via the union bound:
    \begin{align*}
        \underline{q}_{\alpha+} := \quantile{\alpha - \eta}{\mathset{\sLowCdfE(\vx_i, y_i)}_{(\vx_i, y_i)\in\Dcal}} \le \underline{q_\alpha} := \quantile{\alpha - \eta/2}{\mathset{\sLowCdf(\vx_i, y_i)}_{(\vx_i, y_i)\in\Dcal}}
    \end{align*}
    This is because every element $\sLowCdfE(\vx_i, y_i)$ in the first set is lower than the corresponding element in the other set. Now given the new test datapoint $\pertx_{n+1}$, the new calibration scores $\mathset{\sLowCdf(\vx_i, y_i):(\vx_i, y_i)\in\Dcal}$ and $\sLowCdf(\vx_{n+1}, y_{n+1})$ are exchangeable, as a result for the clean corresponding point $\vx_{n+1}$ we have $\prob{\sLowCdf(\vx_{n+1}, y_{n+1})\ge \underline{q_\alpha}} \ge 1 - \alpha + \eta$. Therefore we have the following chain of inequalities:
    \begin{align*}
        \underline{q}_{\alpha+}  \underset{1 - \eta/2}{\le} \underline{q_\alpha}  \underset{1 - \alpha + \eta} {\le} \sLowCdf(\vx_{n+1}, y_{n+1}) \le \smooths(\pertx_{n+1}, y_{n+1})  \underset{1 - \eta/2}{\le} \smooths_+(\vx_{n+1}, y_{n+1})
    \end{align*}
    summing up the probability of each inequality we have $\prob{y_{n+1} \in \conservativeSet_{\alpha+}} = \prob{\smooths_+(\vx_{n+1}, y_{n+1}) \ge \underline{q}_{\alpha+}} \ge 1 - \alpha$
\end{proof}

\begin{proof}[Proof of \autoref{thrm:poisoning-corrected}]
    Since \autoref{eq:poison-problem-mlp} (and \autoref{eq:poison-problem}) is a minimization problem, dropping $a_i \le \sHigh_i$ does not change the optimal solution. For each calibration point, we have:
    \begin{align*}
        s_\mathrm{mc}(\boldsymbol{x} _i, y _i) \underset{1 - \eta/2|\Dcal|}{\ge} \hat{s}(\boldsymbol{x} _i, y _i)-\epsilon _\mathrm{hoef} \ge \underline{s} _\mathrm{cdf}(\tilde{\boldsymbol{x}} _i, y _i) -\epsilon _\mathrm{hoef}
        \underset{1 - \eta/2|\Dcal|}{\ge} \underline{s} _\mathrm{cdf +}(\tilde{\boldsymbol{x}} _i, y _i) -\epsilon _\mathrm{hoef} 
    \end{align*}
    Thus $\underline{s} _\mathrm{cdf+}(\tilde{\boldsymbol{x}}_i, y_i) -\epsilon_\mathrm{hoef} \le s_\mathrm{mc}(\boldsymbol{x} _i, y _i)$ holds with $1 - \eta$ probability for all $i$ via union bound. This follows that $\underline{q}_{\alpha+} \le q_{\alpha, \mathrm{mc}}$ where $q_{\alpha, \mathrm{mc}}$ is the $\alpha$-quantile of MC scores for clean calibration set. Therefore by exchangeability, we have $\prob{s_\mathrm{mc}(\vx_{n+1}, y_{n+1}) \ge q_{\alpha, \mathrm{mc}}} \ge 1 - \alpha = 1 - \alpha' + \eta$. Finally \begin{align*}
        \prob{s_\mathrm{mc}(\vx_{n+1}, y_{n+1}) \ge \underline{q}_{\alpha+}} \underset{1 - \eta}{\ge} \prob{s_\mathrm{mc}(\vx_{n+1}, y_{n+1}) \ge q_{\alpha, \mathrm{mc}}} \ge 1 - \alpha' + \eta
    \end{align*}
    Therefore, $\prob{s_\mathrm{mc}(\vx_{n+1}, y_{n+1}) \ge \underline{q}_{\alpha+}} \ge 1 - \alpha'$.
\end{proof}

\subsection{Details on $l_2$ CDF bounds}
\label{sec:cdf-bounds}

\parbold{Rephrase from \citet{kumar2020certifying}}
The upper bound in \autoref{eq:gaussian_cdf} is a rephrasing of Theorem 2 from \citet{kumar2020certifying}. In the original version the bins are defined as $a < c_1 \le c_2 \le \dots \le c_n < b$. For the for a score function $s$ and (clean) input $\vx$ the statistics $p_{c_j}$ is defined as 
\[ p_{c_j} = \mathrm{Pr}_{\vdelta \sim \gN(\vzero, \sigma^2\mI)}[s(\vx + \vdelta, y) \ge c_i] \]
Here we correct for finite sample estimation via Dvoretzky–Kiefer–Wolfowitz inequality. With the detailed discussion on Monte-Carlo sample correction in \autoref{sec:monte-carlo}, we assume that the statistics are computed with the error correction. For the adversarial point $\pertx \in \gB_r(\vx)$ we have the following upper bound:
\begin{align}
    \hat{s}(\pertx, y) \le c_1 + (b - c_n) \Phi_\sigma\left(\Phi_\sigma^{-1}(p_{c_n}) + r \right) + \sum_{j=1}^{n-1}(c_{j+1} - c_{j}) \Phi_\sigma\left(\Phi_\sigma^{-1}(p_{c_j}) + r \right)
    \label{eq:other-paper-cdf}
\end{align} 

In \autoref{eq:gaussian_cdf} we rewrote the same inequality with a simpler notation. Here we show that the two inequalities are the same. 
Our bins are indexed as $a = b_1 < b_2 \le b_3 \le \dots \le b_{m-1} < b_m = b$; therefore for the same number of bins ($n = m-2$), there is an index mapping as $\forall 1 \le i < m: c_{i-1} = b_i$. Rewriting \autoref{eq:other-paper-cdf} with the new bins, we have:
\begin{align*}
\begin{aligned}
    \hat{s}(\pertx, y) \le &\  b_2 + (b_m - b_{m-1}) \Phi_\sigma\left(\Phi_\sigma^{-1}(p_{b_{m-1}}) + r \right) + \sum_{j=1}^{m-3}(b_{j+2} - b_{j+1}) \Phi_\sigma\left(\Phi_\sigma^{-1}(p_{b_{j+1}}) + r \right) \\
    = &\ b_2 + \sum_{j=1}^{m-2}(b_{j+2} - b_{j+1}) \Phi_\sigma\left(\Phi_\sigma^{-1}(p_{b_{j+1}}) + r \right) = b_2 + \sum_{j=2}^{m-1}(b_{j+1} - b_{j}) \Phi_\sigma\left(\Phi_\sigma^{-1}(p_{b_{j}}) + r \right)
\end{aligned}
\end{align*}

We write the upper bound in terms of CDF function where $p_j = \mathrm{Pr}_{\vdelta \sim \gN(\vzero, \sigma^2\mI)}[s(\vx +\vdelta, y) \le b_j]$. We use two properties from Gaussian distribution \begin{enumerate*}[label=(\roman*)]
    \item for the CDF function it holds that $\Phi_\sigma(-z) = 1 - \Phi_\sigma(z)$
    \item for the quantile (inverse CDF) function it holds that $\Phi^{-1}_\sigma(1-z) = - \Phi^{-1}_\sigma(z)$.
\end{enumerate*} Hence, we have
\begin{align*}
\begin{aligned}
    p_{b_j} =  1 - p_j \Rightarrow \Phi_\sigma\left(\Phi_\sigma^{-1}(p_{b_{j}}) + r \right) &\ =  \Phi_\sigma\left(\Phi_\sigma^{-1}(1 - p_{j}) + r \right) \\
    &\ = \Phi_\sigma\left(-\Phi_\sigma^{-1}(p_{j}) + r \right)\\
    &\ = \Phi_\sigma\left(-[\Phi_\sigma^{-1}(p_{j}) - r] \right)\\
    &\ = 1 - \Phi_\sigma\left(\Phi_\sigma^{-1}(p_{j}) - r \right)\\
\end{aligned}
\end{align*}
It follows
\begin{align*}
    \begin{aligned}
        b_2 + \sum_{j=2}^{m-1}(b_{j+1} - b_{j}) \Phi_\sigma\left(\Phi_\sigma^{-1}(p_{b_{j}}) + r \right) &\ = 
        b_2 + \sum_{j=2}^{m-1}(b_{j+1} - b_{j}) \left[1 - \Phi_\sigma\left(\Phi_\sigma^{-1}(p_{j}) - r \right)\right]\\
        &\ = 
        b_2 + \sum_{j=2}^{m-1}(b_{j+1} - b_{j}) - \sum_{j=2}^{m-1}(b_{j+1} - b_{j}) \Phi_\sigma\left(\Phi_\sigma^{-1}(p_{j}) - r \right)\\
        &\ = 
        b_m - \sum_{j=2}^{m-1}(b_{j+1} - b_{j}) \Phi_\sigma\left(\Phi_\sigma^{-1}(p_{j}) - r \right)
    \end{aligned}
\end{align*}

Intuitively, with a fixed set of bins, the mass of each bin can be bounded within $\gB_r(\vx)$ independently (the bound for each bin is similar to the mean bound). Therefore for a discrete empirical CDF of scores around $\vx$, first we find a worst-case upper bound CDF, then we bound the mean via the Anderson inequality (\autoref{eq:anderson}) given the worst-case CDF.  

\parbold{Lower bounds within $\gB_r(\cdot)$}
Similar to the mean upper bound from the Anderson inequality (\autoref{eq:anderson}), the mean can be lower bounded as:
\begin{align}
    \label{eq:anderson-low}
    \E[h(\vx)] \ge \sum_{j = 2}^{m} b_{j - 1}\cdot[F_h(b_j) - F_h(b_{j - 1})] = b_{m-1} - \sum_{j = 2}^{m-1}F_h(b_j)\cdot(b_{j} - b_{j - 1})
\end{align}
The lower and upper bounds are intuitive as they assume every point within each bin $[b_{j - 1}, b_j)$ is equal to $b_{j - 1}$ for lower and $b_j$ for the upper bound. The rest is just computing the average based on the relative frequency $F_h(b_j) - F_h(b_{j - 1})$. With that the lower bound version of \autoref{eq:certificate-smooth-cdf} is 
\begin{align}
    \label{eq:certificate-lower-cdf}
    \hat{s}(\pertx, y) \ge \sLowCdf(\vx, y) = b_{m-1} - \sum_{j = 2}^{m-1}\Phi_\sigma\left(\Phi_\sigma^{-1}(p_{j}) + r \right)\cdot(b_{j} - b_{j - 1})
\end{align}

Here we derive the equality in \autoref{eq:anderson} -- namely the following;
\begin{align*}
    \sum_{j=2}^{m} b_{j} \cdot [(F_h(b_{j})  - F_h(b_{j-1})]  
        = & \ b_m - \sum_{j=2}^{m-1} F_h(b_{j}) \cdot (b_{j+1} - b_{j})
\end{align*}
The lower bound follows a similar way to derive. We have
\begin{align*}
\begin{aligned}
    \sum_{j=2}^{m} b_{j} &\cdot [(F_h(b_{j})  - F_h(b_{j-1})]  \\
        = &\ b_2 \cdot \left[ (F_h(b_{2})  - F_h(b_{1}) \right] + b_3 \cdot \left[ (F_h(b_{3})  - F_h(b_{2}) \right] + \dots + b_{m} \cdot \left[ (F_h(b_{m})  - F_h(b_{m-1}) \right]\\
    = &\ - b_2\cdot F_h(b_1) + \left[ b_2 \cdot F_h(b_2) - b_3 \cdot F_h(b_2) \right] + \dots + \left[ b_{m-1} \cdot F_h(b_{m-1}) - b_m \cdot F_h(b_{m-1}) \right] + b_m\cdot F_h(b_m)
\end{aligned}
\end{align*}
With $F_h(b_1) = 0$ and $F_h(b_m) = 1$ we have
\begin{align*}
\begin{aligned}
    \sum_{j=2}^{m} b_{j} &\cdot [(F_h(b_{j})  - F_h(b_{j-1})]  \\
    = &\ 0 + \left[-F_h(b_2) \cdot (b_3 - b_2) \right] + \dots + \left[-F_h(b_{m-1}) (b_m - b_{m-1}) \right] + b_m \\
    = &\ b_m - \sum_{j=2}^{m-1} F_h(b_{j}) \cdot (b_{j+1} - b_{j})
\end{aligned}
\end{align*}

\section{Estimating Expectations with Monte-Carlo Sampling}

\label{sec:monte-carlo}
\parbold{Concentration inequalities}
For any random variable $z$, let $z_1, \dots, z_m$ be Monte-Carlo samples of $z$. With $\E_{m}[z] = \frac{1}{m}\sum_{i = 1}^m z_i$, we bound the true expectation around the MC-estimate via Hoeffding's inequality. The following holds with any adjustable $1 - \eta$ probability;
\begin{align*}
    |\E[z] - \E_m[z]| \le \sqrt{\frac{\log(\frac{2}{\eta})}{2m}}
\end{align*}
This bound only accesses to the empirical mean and not the samples. Therefore, in cases where we want to account for the distance of empirical mean, and the true expectation for an unknown variable, we can use this bound. An example of this case is the test-time correction where the upper bound on the mean of the unseen point is computed while the empirical mean is not computable (since there are no samples). 

Let $\sigma^2_m$ be the variance of the MC samples, then empirical Bernstein inequality produces variance-dependent confidence intervals as following:
\begin{align*}
    |\E[z] - \E_m[z]| \le \sqrt{2\sigma^2_m\frac{\ln(\frac{4}{\eta})}{m}} + \frac{7\ln(\frac{4}{\eta})}{3(m - 1)}
\end{align*}
Similar to the mean, the empirical CDF is also bounded between an upper and a lower CDF, via the Dvoretzky–Kiefer–Wolfowitz (DKW) inequality \citep{dvoretzky1956asymptotic}. Let $F(b_i) = \prob{z \le b_i}$ and $F_m(b_i) = \sum_{j=1}^m \1[z_j \le b_i]$,
\begin{align*}
    |F(b_i) - F_m(b_i)| \le \sqrt{\frac{\log(\frac{2}{\eta})}{2m}}
\end{align*}
The above inequality holds simultaneously for all $b_i$.

For \autoref{eq:certificate-smooth} we use the Bernstein inequality as is has shown a better empirical result compare to Hoeffding's inequality. For \autoref{eq:certificate-smooth-cdf} we use the DKW inequality to find confidence intervals the empirical CDF.

\parbold{Error correction in \autoref{eq:certificate-smooth} and \autoref{eq:certificate-smooth-cdf}}
To find the upper (or lower) bound in \autoref{eq:certificate-smooth}, we need to estimate the mean of the smooth score around the input $\vx$. We use the mean corrected with the Bernstein confidence interval. For the upper bound problem, we use the upper end of the interval since it is more conservative. The same logic follows for the lower bound.

\begin{wrapfigure}{r}{0.45\textwidth}
    \centering
    \input{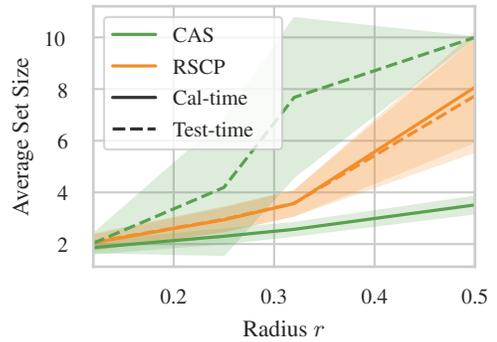}
    \caption{Comparison of \method and RSCP for faster (calibration-time) and test-time error correction.}
    \label{fig:err-both}
\end{wrapfigure}

For \autoref{eq:certificate-smooth-cdf} we use the Dvoretzky–Kiefer–Wolfowitz inequality to find an upper (or lower) CDF. Since in the \autoref{eq:anderson} the CDF is added with a negative sign, the lower endpoint of the confidence interval should be used to find a conservative upper bound.

Empirically, Bernstein's confidence intervals are tighter than Hoeffding's intervals. Therefore we only use the Hoeffding error anytime we need a correction without having access to the variance.

\parbold{Test-time correction \citep{anonymous2023provably}} 
The MC-sampled smooth score does not break the exchangeability since this estimation is permutation invariant. This means that given the clean input $\vx_{n+1}$ the estimated scores are exchangeable and the guarantee is valid without any error correction. However, given $\pertx$, \method and RSCP find bounds on the true mean. Given $\pertx$, we compute $\sHighE$ via solving either \autoref{eq:certificate-smooth} (RSCP) or \autoref{eq:certificate-smooth-cdf} (\method) with the error corrected estimate. For both methods, the following holds:
\begin{align*}
    q_{\alpha, \mathrm{mc}} \underset{1 - \alpha}{\le} \smooths_{mc}(\vx_{n+1}, y_{n+1}) \underset{1 - \eta_1}{\le} \smooths(\vx_{n+1}, y_{n+1}) + \epsilon_\mathrm{hoef} \le \sHigh(\pertx_{n+1}, y_{n+1}) + \epsilon_\mathrm{hoef} \underset{1 - \eta_2}{\le} \sHighE(\pertx_{n+1}, y_{n+1}) + \epsilon_\mathrm{hoef}
\end{align*}
By setting $\alpha' = \alpha + \eta_1 + \eta_2$ we have a valid CP guarantee with certified $1 - \alpha$ probability. 

\parbold{Calibration-time vs test-time correction}
As shown in \autoref{fig:err-both}, \method benefits significantly from calibration-time robustness. 
The reason is that the CDF bound (\autoref{eq:certificate-smooth-cdf}) performs significantly better than the mean bound (\autoref{eq:certificate-smooth}) when the score distribution is more spread out. For distributions concentrated around each endpoint of the domain, the CDF has a high slope at the endpoint and is almost flat elsewhere. While using DKW inequality, a large penalty is added to the distribution resulting in larger CDF intervals. Meanwhile, in these distributions, the mean bound can benefit from Bernstein inequality which due to the low variance performs even better.  In calibration-time robustness, we find the lower bound for true scores (which are often more spread) while in test-time unlikely classes that have scores concentrated to 0 are bounded by large value (due to DKW for concentrated scores) which directly affects the set size. In addition, \citet{anonymous2023provably} adds a Hoeffding error to the unseen clean score, where in our method we bound the estimation of the input which can use the Bernstein error.
Since we are free to choose between test-time and calibration-time correction, and RSCP has equal performance for both, we argue that we should use calibration-time correction as a default.
For \autoref{fig:three-plots-evasion} we choose the best performance of each method in either calibration- or test-time robustness with error correction.

\section{Details on RSCP}
\label{sec:details-rscp}
\subsection{Equivalence Between RSCP and our Gaussian Baseline Bound}
\label{sec:comparison-rscp}

For a given score function $s:\gX\times\gY \mapsto[0, 1]$ on continuous inputs, \citet{gendler2021adversarially} define the new scoring function as follows:
\begin{align}
    \label{eq:rscp-score}
    s_{\mathrm{rscp}}(x, y) = \Phi^{-1}(\E[s(\smoothf(x), y)])
\end{align}
RSCP computes the $\alpha$-quantile $q_\alpha$ of the new calibration scores (\autoref{eq:rscp-score}) and compares each score with the modified threshold\footnote{
Originally RSCP shifts the quantile forward $\underline{q_\alpha} = q_\alpha + r/\sigma$ since it is defined with the non-conformity setup where scores lower than the quantile are accepted. Here since we use conformity (agreement) scores and the acceptance criteria is to be larger than $q_\alpha$ we shift the quantile backward. The setups are equivalent via changing the sign of the scores (see \autoref{sec:conformal-more}).} $\underline{q_\alpha} = q_\alpha - r/\sigma$
, where $r$ is the radius of the $l_2$ ball from the threat model, and $\sigma$ is the scale of the smoothing distribution. We can equivalently add an additional $r/\sigma$ term to test scores instead and compare the augmented score with unchanged $q_\alpha$. Using $\Phi^{-1}_\sigma(p) = \sigma\Phi^{-1}(p)$ as a property of the inverse CDF function of the Gaussian distribution, we have
\begin{align*}
    \Phi^{-1}\left(\E[s(\smoothf_\sigma(\vx), y)] \right) \le \Phi^{-1}\left(\E[s(\smoothf_\sigma(\pertx), y)] \right) + \frac{r}{\sigma} \Rightarrow \Phi^{-1}_\sigma\left(\E[s(\smoothf_\sigma(\vx), y)] \right) \le \Phi^{-1}_\sigma\left(\E[s(\smoothf_\sigma(\pertx), y)] \right) + r \\
\end{align*}
Since the CDF is a monotonically increasing function we apply $\Phi_\sigma$ on both sides of the inequality:
\begin{align*}
    \Phi_\sigma\left(\Phi^{-1}_\sigma\left(\E[s(\smoothf_\sigma(\vx), y)] \right)\right) \le \Phi_\sigma\left(\Phi^{-1}_\sigma\left(\E[s(\smoothf_\sigma(\pertx), y)] \right) + r\right)\\
    \Rightarrow \E[s(\smoothf_\sigma(\pertx), y) \le \Phi_\sigma\left(\Phi^{-1}_\sigma\left(\E[s(\smoothf_\sigma(\pertx), y)] \right) + r\right)
\end{align*}
Substituting $p=\E[s(\smoothf(\pertx), y)]$ we see that 
this is equivalent to the Gaussian $\sHighMean$ upper-bound defined in \autoref{sec:efficienct-bounds}. 


\subsection{Comparison with \citet{cauchois2020robust}}
\label{sec:comparison-other-robust}
\citet{cauchois2020robust} derive robust prediction sets when the $f$-divergence between the test distribution and the calibration distribution of the non-conformity scores is bounded by a fixed value $\rho$. 
We can connect their approach to our definition of adversarial robustness using the results from \citet{Dvijotham2020AFF}.
Specifically, we can rewrite the optimization problem $\max_{\pertx \in \gB_r(\vx)} \E[s(\smoothf(\pertx, y)]$ over the ball $\gB_r(\vx)$ to the optimization problem
$
\max_{\nu \in \gP} \E[s(\nu(\vx, y)]
$
over the space of probability measures 
$\gP = \{ \smoothf(\pertx) \mid \vx \in \gB_r(\vx) \}$. Since this set is intractable we can relax the problem using the fact that 
$
\gP \subseteq \{ D_f (\nu || \smoothf) \le \rho_{r}^f  \} 
$
for an appropriately chosen $\rho_{r}^f$ where  $D_f (\nu || \smoothf)$ is the $f$-divergence between the smoothing distribution $\nu$ centered at a perturbed example and the smoothing distribution $\smoothf$ centered at the clean example. See \citet{Dvijotham2020AFF} for a derivation of the optimal $\rho_{r}^f$ for different different divergence functions $f$ and different smoothing distributions. 
Thus, for smooth scores there is a direct connection between RSCP, \method and \citet{cauchois2020robust}'s method.

Importantly however, for most choices of $f$ (e.g. the KL divergence) the relaxation results in a looser (though potentially easier to compute) bound. 
The analysis in \citet{Dvijotham2020AFF} was developed for classification problems but it also directly applies to our setting. They show that we need to use the Hockey-
Stick divergences with the right parameters to obtain tight certificates. Specifically, for Gaussian smoothing and an $l_2$ norm the result is equivalent to the tight certificate from \citet{cohen2019certified}.
Disregarding that Hockey-Stick divergences are harder
to estimate in general, it means that in the best case, the approach by \citet{cauchois2020robust} can recover the baseline $\sHighMean(\pertx, y)$ which we have shown is looser than our $\sHighCdf(\pertx, y)$.

\section{Technical Details on Poisoning Certificate}
\label{sec:technical-poisoning}

\parbold{Feature poisoning} The solution of the optimization problem in \autoref{eq:poison-problem} is robust to feature poisoning; however, the problem is hard to solve since: \begin{enumerate*}[label=(\roman*)]
    \item we need to optimize over each $\vz_i$ in $\gB(\pertx_i)$,
    \item it involves a quantile computation,
    \item and it has a cardinality constraint as the sum of indicator functions.
\end{enumerate*} Therefore, we relax the problem to a MILP which can solve with standard solvers.
First, we replace each $\vz_i \in \gB(\pertx_i)$ constraint with a $\sLow_i \le \tilde{s}_i \le \sHigh_i$ constraint
directly over scores $s_i$ where the lower and upper bounds are computed as in discussed in \autoref{sec:efficienct-bounds}. This is a sound relaxation and the optimal $\underline{q_\alpha}$ of the relaxed problem is smaller or equal than the $\underline{q_\alpha}$ of the original problem.
Then, we introduce $|\Dcal|$ binary variables to compute the $\alpha$ quantile, and additional $|\Dcal|$ binary variables to enforce the perturbation budget. The resulting MILP is:
\begin{equation}
    \label{eq:poison-problem-mlp}
    \begin{aligned}
        \underline{q_\alpha} = \quad  &\min_{s_i, q} \quad  q \\
        s.t. \quad  & \forall \perts_i: \sLow_i \le s_i \le \sHigh_i \\
        \quad &  t_i := \1[s_i \le q], \quad \sum_{i = 1}^n z_i \le \mathfloor{\alpha n}, \quad \mathrm{ and } \quad \sum_{i = 1}^n (1 - t_i) \le \mathceil{(1 - \alpha) n} \\
        \quad &  b_i := \1[s_i \neq \perts_i], \quad \sum_{i = 1}^n b_i \le k \\
    \end{aligned}
\end{equation}

In \autoref{eq:poison-problem-mlp}, the $z_i$ variables indicate whether the calibration point is below or above the $\alpha$ quantile $q$, and the $b_i$ variables indicate whether the point is perturbed or not. We use the standard big-M technique to translate this into a canonical form which we solve with MOSEK.

\parbold{Label poisoning} We can directly rewrite \autoref{eq:label-poison-problem} as a MILP without any relaxations. Let $\mS$ be an $n=|\Dcal|$ by $c$ matrix of scores for each class and each calibration point, where $c$ is the number of classes.
We have

\begin{equation}
    \label{eq:label-poison-problem-mlp}
    \begin{aligned}
        \underline{q_\alpha} = \quad  &\min_{q, \mC \in \{0, 1\}^{n \times c}} \quad  q \\
        \text{s.t.} \qquad    & \mC \1^{c \times 1} = \1^{n \times 1} \\
        & \vr = (\mC \odot \mS)\times \1^{c \times 1}  \\
         &\sum_i \mC[i, y_i] \geq n-k \\
        \quad \quad & z_i := \1[r_i \le q], \quad \sum_{i = 1}^n z_i \le \mathfloor{\alpha n}, \quad \mathrm{ and } \quad \sum_{i = 1}^n (1 - z_i) 
        \le \mathceil{(1 - \alpha) n} 
    \end{aligned}
\end{equation}
where the binary one-hot matrix $\mC$ is responsible for selecting one score per calibration point (i.e. one of the $c$ possible labels), $\vr$ is the resulting set of chosen scores, and the $z_i$ variables implement the quantile as before. 

\parbold{Complexity}
Note that while in general, solving MILPs is computationally expensive, since our calibration sets are relatively small, we can still obtain the exact solution in reasonable wall-clock time. We leave it as future work to derive more efficient algorithms for the feature and label poisoning problems.

\section{Robustness to Poisoning and Evasion Attacks Combined}
\label{sec:combined-robustness}

In \autoref{sec:robust-to-poisoning} we make CP robust to poisonings (in feature or label domain) by finding a conservative $\hat{q}$ that in the most adverse case of attack (within the defined budget and threat model) the coverage probability remains above $1 - \alpha$. Once this threshold is defined, we can consider the calibrated quantile to safely satisfy the guarantee on the clean test -- we can assume that CP was calibrated on clean calibration data. Formally the solution to \autoref{eq:poison-problem}, and \autoref{eq:label-poison-problem}, is a threshold with which the prediction sets constructed for clean $\vx$ has larger than $1 - \alpha$ coverage probability.

While making CP robust to evasion, we only consider the confidence interval of scores for the clean test point given the potentially perturbed point $\pertx$. This process only involves computing upper bounds on the given test point and hence is independent of the prior robustness to poisoning. In other words, the resulting conservative prediction set includes the prediction set of the clean datapoint $\gC(\vx) \subseteq\conservativeSet(\pertx)$. 

This shows that we can make CP robust to poisoning and evasion attacks at the same time. However, this combined robustness comes at the price of comparably larger prediction sets. The robust $\underline{q}$ is less than $q_\alpha$ which allows more labels to be included in the prediction sets. At the same time, for each test point, the upper-bound scores introduce a higher probability for a label to be included in the prediction sets again. So there will be two conservative processes each increasing the chance of accepting a label which increases the expected set size.

\section{Time and Space Complexity}
\label{sec:time-and-space}

Our robust CP approach breaks down into four computations \begin{enumerate*}[label=(\roman*)]
    \item computing the score function, 
    \item estimating expectations for randomized smoothing (in practice the MC sampling and computing the confidence intervals),
    \item computing upper-bounds, and
    \item standard CP processes including calibration and constructing prediction sets.
\end{enumerate*}
Here we omit the time complexity analysis of the model, and with the black-box access, we assume the model's prediction of logits to take $\gO(1)$ step. The computation of the conformity score depends on the choice of this function. TPS takes $\gO(K)$ ($K$ is the number of classes) to compute the categorical distribution via softmax function. APS score function takes an additional $\gO(K)$ steps to sort the class probabilities and compute the summation of confidences (see \autoref{sec:background} for the definition of the score function). This additional sort can become time-consuming for datasets with large number of classes (like \texttt{ImageNet}) For simplicity, we call the score function to take $t_s$ steps. Standard CP procedures are calibration and constructing prediction sets. Given $n$ calibration score finding the $1 - \alpha$ quantile takes $\gO(n)$ steps (median computation) and the prediction sets take $\gO(C)$ to be constructed for each test input. All the time complexities are reported w.r.t. serial computation, while with enough number of parallel processing cores, all above computations can be done in relatively lower number of steps. 

In the randomized smoothing we need to estimate the expected score function within the smoothing scheme. For that, we use Monte-Carlo sampling which takes $\gO(N\times M)$ steps to compute the mean of $M$ Monte-Carlo samples and $N$ is the number of datapoints in total.  

With the Monte-Carlo samples each upper- and lower-bound need solving an optimization problem. The optimal value is found via a closed-form solution for Gaussian smoothing. Given $S$ bins for the binary (and discrete) CDF computing this bound takes $\gO(S \times R)$ time where $R$ is the number of regions of similar likelihood and we have $R = r_a + r_d + 1$. We refer to the time computation time of the bound as $t_b$.

\begin{figure}[]
    \centering
    \input{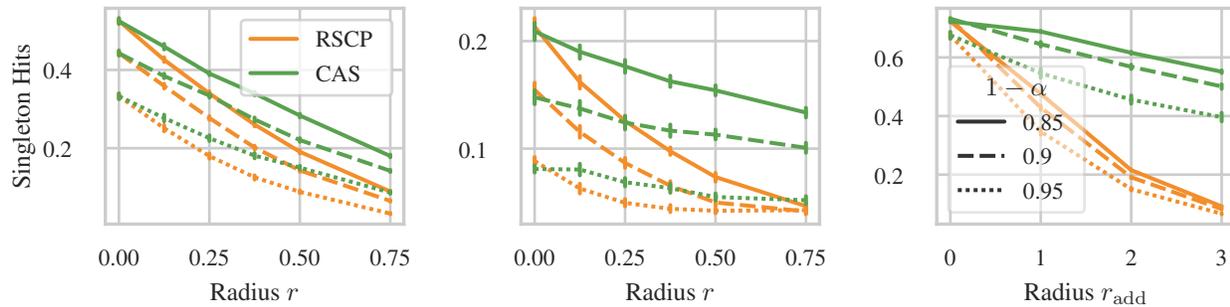}
    \caption{Singleton hit ratio of \method and RSCP under evasion for (from left to right) \texttt{CIFAR-10} with APS, \texttt{ImageNet} with TPS, and \texttt{Cora} with APS.}
    \label{fig:singleton-hits}
\end{figure}

As a result, in the evasion setup, we take $\gO(NM)$ additional steps for calibration on smooth scores and $\gO(MK + K \cdot t_b)$ for constructing the prediction sets. We also proposed a faster way to provide robust prediction sets in \autoref{sec:calibration-robustness}. For that we compute an upperbound per each calibration datapoint but only for the true class. For any given test point we only compute smooth scores which in total reduces the computation to $\gO(MK)$ for test time (per test datapoint), and increases the calibration time complexity to $\gO(N\cdot t_b)$. This procedure decreases the number of steps in total. \autoref{tab:comparison_runtime} compares the runtime of both approaches for a limited number of calibration, and test points.

For poisoning in the feature space, we should first compute the upper and lower bounds for each calibration data which takes $\gO(NM + M\cdot t_b)$ steps. Here we just compute the bounds for the true label. We then solve a mixed integer linear programming which is computationally hard. We apply tricks like big-M method to make the problem solvable and enable the use of standard convex optimization solvers. Similarly for the label poisoning, the problem is hard involving ILP solvers, but here we do not need to compute bounds on scores as the perturbations are in the label domain.

\section{Supplementary To Experiments}
\label{sec:additional-experiments}

\subsection{Details on the Experiments in the Manuscript}

In our core experiment, we utilized a \texttt{ResNet-110} model pre-trained on the \texttt{CIFAR-10} dataset and a \texttt{ResNet-50} model pre-trained on the \texttt{ImageNet} dataset. Both models were trained using noisy training by Gaussian data augmentation across various noise variances, as proposed by \citet{lecuyer2019certified} and later used by \citet{cohen2019certified} for randomized smoothing. Detailed insights into the model training and augmentation processes are elaborated in \citet{cohen2019certified, salman2019provably}.

For evaluation, we employed an $l_2$ norm smoothing paradigm and applied various noise levels, identifying the model that delivered optimal performance based on findings from \citet{cohen2019certified}. 
On \texttt{CIFAR-10} dataset we used a skip parameter and ran the experiments on between 1000 to 2000 samples. 
Similarly, 500 data points are used from the sampling of every 100-image from the \texttt{ImageNet} dataset. Noise variance settings used were $\sigma=0.25$ for \texttt{CIFAR-10} and $\sigma=0.5$ for \texttt{ImageNet}. During the Monte Carlo sampling, each datapoint was processed through $10^4$ iterations to calculate the expected probability or mean. 

For our experiments on the \texttt{Cora-ML} dataset. we utilized a two-layer \texttt{GCN} equipped with 64 hidden units. Followed by \citet{bojchevski2020efficient}, our training procedure incorporated randomized perturbations of the node features. Specifically, we used a perturbation addition probability $(p_{+})$ of $0.01$ and a deletion probability $(p_{-})$ of $0.6$. For the training process, we employed 20 node labels per class for training and similar number of nodes for validation. We conducted the training over 1,000 epochs. The remaining portion of the dataset was set aside for evaluation purposes.

In our conformal prediction strategy, the split conformal method was adopted. To account for the effect of randomness in calibration set sampling, we reported our result in terms of mean and confidence bounds over 100 calibration samplings.

Moreover, results for adversarial cases are discussed. For these attacks, we employed the projected gradient descent (PGD) attack \citep{madry2017towards}, using an alpha value of 0.1 across 40 iterations. The attack outcomes, constrained by L2 norm distance from the original image, are presented for $r= 0.125$.


\parbold{Singleton hits ratio} This metric quantifies the proportion of correct singleton predictions which can be used without any further post-processing. Similar to the prediction set size, \autoref{fig:singleton-hits} shows that \method outperform RSCP on all datasets.

\begin{figure}[t]
    \centering
    \input{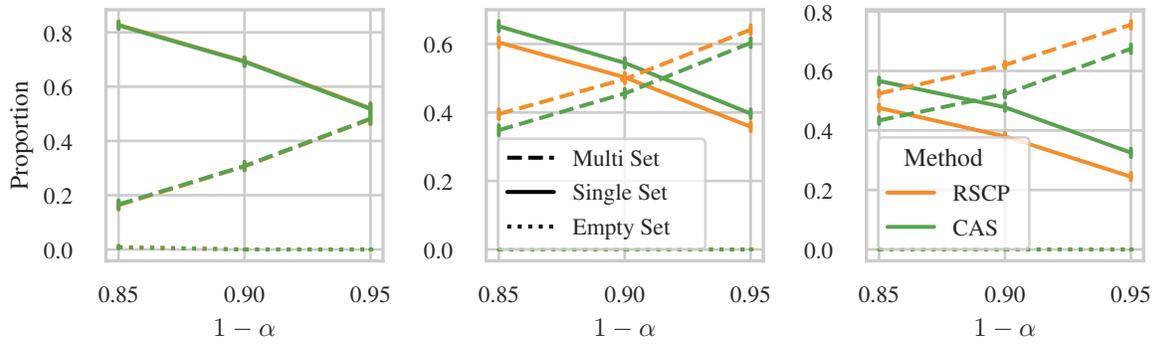}
    \caption{The proportion of singleton, empty and multi-sets for RSCP and \method across radii (left) $r = 0$, (middle) $r=0.12$, and (right) $r=0.25$.}
    \label{fig:zero-one-mutli}
\end{figure}

\parbold{Proportion of Empty, Singleton, and Multi-sets}
While we report the average set size (similar to many other studies in CP), a CP method might misleadingly show to be more efficient by returning more empty sets. That is why an alternative metric is to only report the average size of non-empty sets. In \autoref{fig:zero-one-mutli} we report the proportion of empty, singleton and multi-prediction sets for various radii.  In vanilla CP, as we increase the $1 - \alpha$ guarantee to higher values, CP adds more elements to prediction sets to satisfy the increased coverage guarantee. Since there are almost no empty prediction sets for various $\alpha$, and $r$, both effective and average set size are the same. 

\begin{figure}[]
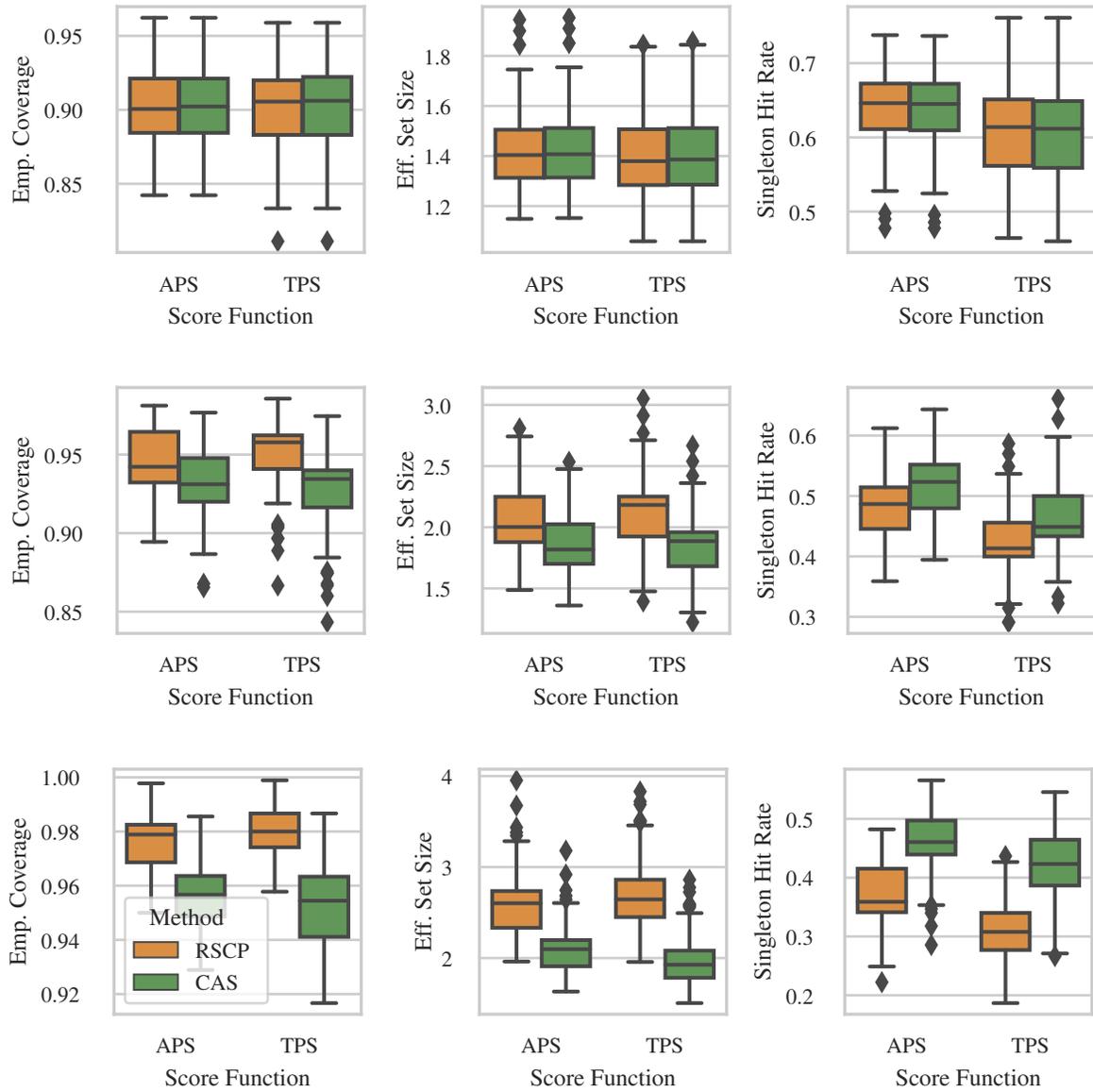

    \centering
    \input{imgs/tps-aps-bounds-results-radi000-sigma0025.pgf}\\
    \input{imgs/tps-aps-bounds-results-radi0012-sigma0025.pgf}\\
    \input{imgs/tps-aps-bounds-results-radi0025-sigma0025.pgf}
    \caption{Comparison of RSCP and \method for smooth APS and TPS score across various radii for (left column) empirical coverage (middle column) set size, and (right column) singleton hits. From upper to lower row results are respectively for $r=0$, $r=0.12$, and $r=0.25$. All results are for \texttt{CIFAR-10}, and smoothing with $\sigma=0.25$}
    \label{fig:tps-aps}
\end{figure}

\parbold{Different Score Functions}
As mentioned in \autoref{sec:background} (and in \autoref{sec:conformal-more} extensively), coverage guarantee in vanilla CP, and robustness methods defined on top (including RSCP, and \method) are defined agnostic to the score function leaving the freedom of choosing the score based on the domain of application. Here we empirically support this argument. \autoref{fig:tps-aps} compares RSCP with \method applied on  TPS and APS score functions. In all scores, and all metrics \method shows an improved result.

\newpage
\section{Notations and Definition Guide}
\label{sec:notations}
For a complete guide to all notations used in the paper see \autoref{tab:notations}.

\begin{table}[H]
\centering
\begin{tabular}{@{}rp{0.7\textwidth}@{}}
\toprule
Notation & Desciption                                                                             \\ \midrule
$\vx$        
&
The clean input \\

$(\vx_i, y_i)$        
&
The clean input alongside its true label. \\

$\Dcal$        
&
Clean calibration set. A set of labeled datapoints which its labels are unseen by the model during the training. Precisely, the conformity score (e.g. model softmax) is exchangeable between elements of this set and the test set. \\

$\tilde\vx$        
&
The input perturbed by the adversary \\

$(\vx_i, \tilde{y}_i)$        
&
The clean input alongside a label that is potentially flipped by the adversary. \\

$\tilde{\gD}_\mathrm{cal}$        
&
The poisoned calibration set. Here the adversary has returned a set, given clean $\Dcal$, where under threat model either features are perturbed, or labels are fliped (or both). \\

$\gB(\cdot)$        
&
Point-level threat model: The set of all allowed perturbations w.r.t. the clean point; e.g. all points that are closer than $r$ in $l_2$ distance  \\

$\setgB_{k, \gB}(\gD)$
&
Set-level threat model: The set of all allowed perturbations changing an input set; e.g. CP's calibration set. As an example the set of all perturbed sets where the adversary has changed at most $k$ points within a point-level threat model. \\

$s(\cdot, \cdot)$        
& 
Conformity score function originally defined for vanilla CP\\

$s_\mathrm{rscp}(\cdot, \cdot)$        
& 
Scores defined by \citet{gendler2021adversarially}. \\

$\sHigh(\cdot, \cdot), \sLow(\cdot, \cdot)$        
& 
Upper- and lower-bounds for given score function $s(\cdot, \cdot)$ within the specified threat model. \\

$q_\alpha$        
& 
Conformal quantile computed by CP on the clean calibration set with nominal coverage probability $1 - \alpha$
\\

$\pertq_\alpha$
& 
Adversarial conformal quantile; this is a quantile of the calibration set that is poisoned by the adversary. It is expected that this quantile results in lower coverage compared to $q_\alpha$.
\\

$\underline{q_\alpha}$        
& 
Conservative lower-bound for $q_\alpha$; This is computed by the defender to return robustness prediction sets.
\\

$\gC_\alpha(\cdot)$        
& 
Prediction set of vanilla CP with $1 - \alpha$ nominal coverage. \\

$\conservativeSet_\alpha(\cdot)$      
& 
Prediction set of robust CP with $1 - \alpha$ nominal coverage. Dependent on the attack scenario (evasion or poisoning), this set is robust to the perturbations within the threat model. \\

$\hat{s}(\cdot, \cdot)$      
& 
The smooth score for the input $\vx$. This score is the expectation of the score under a predefined randomized smoothing framework. 
\\

$\sHighMean(\cdot, \cdot)$      
& 
The upperbound score calculated by solving \autoref{eq:certificate-smooth}. This problem only has the mean similarity constraint. 
\\

$\sHighCdf(\cdot, \cdot)$      
& 
The upperbound score calculated by solving \autoref{eq:certificate-smooth-cdf}. This problem only has the CDF similarity constraint. 
\\

\bottomrule
\end{tabular}
\caption{Table of notations used in the paper.}
\label{tab:notations}
\end{table}

\end{document}